\def\year{2022}\relax
\documentclass[letterpaper]{article} 
\usepackage{aaai22}  
\usepackage{times}  
\usepackage{helvet}  
\usepackage{courier}  
\usepackage[hyphens]{url}  
\usepackage{graphicx} 
\usepackage{natbib}  
\usepackage[small]{caption} 
\DeclareCaptionStyle{ruled}{labelfont=normalfont,labelsep=colon,strut=off} 
\usepackage{algorithm}
\usepackage{algorithmic}
\usepackage{nicefrac}

\usepackage{amsmath}
\usepackage{amssymb}
\usepackage[small]{subfigure}
\usepackage{bbm}

\usepackage{blkarray, bigstrut}
\usepackage{enumitem}
\setlist[enumerate,itemize]{itemsep=0.5pt,topsep=0pt}

\usepackage[textsize=small,textwidth=3cm]{todonotes}
\definecolor{kentuckyblue}{RGB}{0, 93, 170}

\newtheorem{theorem}{Theorem}
\newtheorem{definition}{Definition}

\newcommand{\reals}{\ensuremath{\mathbb{R}}}
\newcommand{\mdp}{\ensuremath{\mathcal{M}}}
\newcommand{\nominalmdp}{\ensuremath{\mathcal{M}^{\mathcal{N}}}}
\newcommand{\trueconstrainedmdp}{\ensuremath{\mathcal{M}^{\mathcal{C}^*}}}
\newcommand{\learnedmdp}{\ensuremath{\mathcal{M}^{\mathcal{C}}}}
\newcommand{\states}{\ensuremath{\mathcal{S}}}
\newcommand{\actions}{\ensuremath{\mathcal{A}}}
\newcommand{\demonstrations}{\ensuremath{\mathcal{D}}}
\newcommand{\constraints}{\ensuremath{\mathcal{C}}}

\newcommand{\optnom}{\ensuremath{\pi_{\mathcal{N}}^*}}
\newcommand{\optconstrained}{\ensuremath{\pi_{\mathcal{C}^*}^*}}

\setlist[enumerate,itemize]{itemsep=0.5pt,topsep=0pt}
\renewcommand{\citet}[1]{\citeauthor{#1}~\shortcite{#1}}

\title{Making Human-Like Trade-offs in Constrained Environments by Learning from Demonstrations}

\author{%
    Arie Glazier\textsuperscript{\rm 1 } 
    Andrea Loreggia\textsuperscript{\rm 2 } 
    Nicholas Mattei\textsuperscript{\rm 1 } \\
    Taher Rahgooy\textsuperscript{\rm 3 }
    Francesca Rossi\textsuperscript{\rm 3 }
    K. Brent Venable\textsuperscript{\rm 3,5 }
}
\affiliations {
    \textsuperscript{\rm 1} Tulane University, New Orleans, LA, USA \\
    \textsuperscript{\rm 2} European University Institute, Italy \\
    \textsuperscript{\rm 3} University of West Florida, Pensacola, FL, USA \\
    \textsuperscript{\rm 4} IBM Research, Yorktown Heights, NY, USA \\
    \textsuperscript{\rm 5} Institute for Human and Machine Cognition, Pensacola, FL, USA \\
    \texttt{adglazier@gmail.com, andrea.loreggia@gmail.com, nsmattei@tulane.edu \\ trahgooy@students.uwf.edu, Francesca.Rossi2@ibm.com, bvenable@ihmc.org}
}

\begin{document}

\maketitle

\begin{abstract}
Many real-life scenarios require humans to make difficult trade-offs: do we always follow all the traffic rules or do we violate the speed limit in an emergency? These scenarios force us to evaluate the trade-off between collective norms and our own personal objectives.
To create effective AI-human teams, we must equip
AI agents with a model of how humans make trade-offs in complex, constrained environments. These agents will be able to mirror human behavior or to draw human attention to situations where decision making could be improved.
To this end, we propose a novel inverse reinforcement learning (IRL) method for learning implicit hard and soft constraints from demonstrations, enabling agents to quickly adapt to new settings. In addition, learning soft constraints over states, actions, and state features allows agents to transfer this knowledge to new domains that share similar aspects.
%
We then use the constraint learning method to implement a novel system architecture that leverages a cognitive model of human decision making, multi-alternative decision field theory (MDFT), to orchestrate competing objectives. We evaluate the resulting agent on trajectory length, number of violated constraints, and total reward, demonstrating that our agent architecture is both general and achieves strong performance. Thus we are able to capture and replicate human-like trade-offs from demonstrations in environments when constraints are not explicit.
\end{abstract}

\section{Introduction}

Implicit and explicit constraints are present in many decision making scenarios, and they force us to make difficult decisions: do we always satisfy all constraints, or do we violate some of them in exceptional circumstances? Many techniques can be used to combine constraints and goals so that the agent rationally minimizes constraint violations while achieving the given goal \cite{noothigattu-2019-ethicalvalues}. However, it is well known that humans are not rational. When we need to make a decision in a constrained environment, we often reason by employing heuristics and approximations which are subject to bias and noise \cite{busemeyer2002survey,booch-2021-thinkfast}. This means that optimal techniques may not be suitable if the aim is to design autonomous artificial agents that act like humans, or decision support systems that simulate human behavior to anticipate it and possibly alert humans by making them aware of their reasoning and inference deficiencies.

Moreover, these constraints are often not explicitly given, but need to be inferred from observations of how other agents act in the constrained world. 
Learning constraints from demonstrations is an important topic in the domains of inverse reinforcement learning \cite{scobee-2020-maximumlikelihood,abbeel2004apprenticeship}, which is used to implement AI safety goals including value alignment \cite{russell2015research} and to circumvent reward hacking \cite{amodei2016concrete,ray2019benchmarking}. Recent work has focused on building ethically bounded agents \cite{svegliato2021ethically,FoMa19a} that comply with ethical or moral theories of action. Following the work of \citet{scobee-2020-maximumlikelihood}, we propose an architecture that, given access to a model of the environment and to demonstrations of constrained behavior, is able to learn constraints associated with states, actions, or state features. Our method, MESC-IRL, performs comparably with the state of the art and is more general, as it can handle both hard and soft constraints in both deterministic and non-deterministic environments. It is also decomposable into features of the environment, supporting the transfer of learned constraints between environments.

Once the constraints are learned, we turn our attention to \emph{making human like trade-offs}. Enabling agents to make trade-offs like humans allows them to mirror human behavior or to draw human attention to situations where decision making could be improved \cite{booch-2021-thinkfast}. Additionally, making these trade-offs explicit enables decision support tools that are able to mirror the goals of human decision makers \cite{BaBoMaRo18,balakrishnan2018incorporating}. We propose a novel orchestration technique leveraging Multi-Alternative Decision Field Theory (MDFT) \cite{busemeyer2002survey}, a decision making framework which is based on a psychological theory of how humans make decisions that is able to capture deviations from rationality observed in humans, making trade-offs between competing objectives in a more human-like way. We compare this MDFT-based orchestrator with other methods both theoretically and empirically, showing that our architecture is theoretically more expressive and obtains better empirical performance across a range of metrics when acting in constrained environments. The goal here is to use a cognitive model to capture the sometimes irrational decisions made by humans. Building machines that \emph{act} more like humans is a step to create effective human-machine teams or decision support systems.

%

\section{Preliminaries and Related Work}\label{sec:prelim}

\begin{figure}[t]
  \centering
  \includegraphics[width=0.9\linewidth]{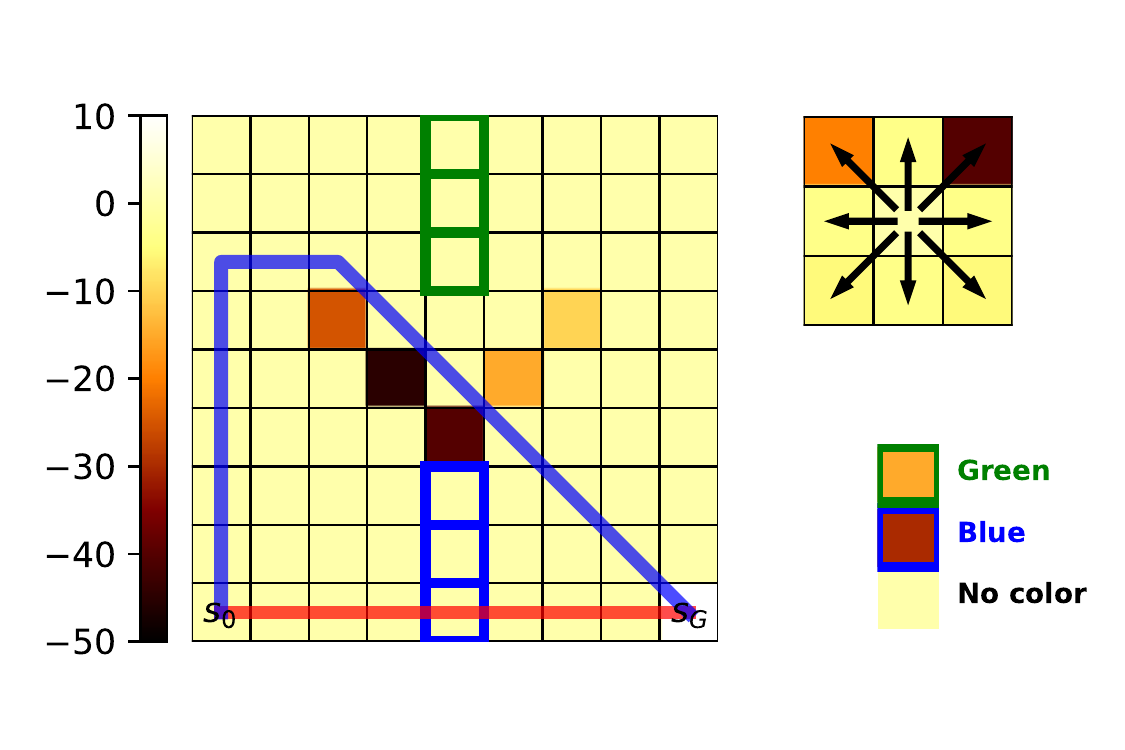}
  \caption{Grid with constraints of varying costs over actions (top right), state occupancy (grid, left) and state features (bottom right). Note that $\optnom$ for $\nominalmdp$ is the red trajectory, obtained by an agent that does not know the constraints while $\optconstrained$ for $\trueconstrainedmdp$ is the blue trajectory.}
  \label{fig:example_grid}
\end{figure}

We begin this section by providing the preliminary notions on the context of our work, that is, constrained Markov Decision Processes and Reinforcement Learning \cite{sutton-barto-rl}. We then review fundamental concepts and methods on Inverse Reinforcement Learning \cite{Ng2000,abbeel2004apprenticeship} and background on Constrained Markov Decision Processes \cite{altman-1999-constrained} including related work on learning constraints \cite{ziebart-2008-maxentropy,anwar-2021-icrl,scobee-2020-maximumlikelihood}, which we will leverage to develop our novel method for learning soft constraints \cite{rossi2006handbook} from demonstrations \cite{chou2018learning}. We conclude this section presenting a short review of the Multi-Alternative Decision Field Theory \cite{busemeyer2002survey}, the cognitive model of decision making which will be at the core of our novel approach to orchestrating competing objectives. 

\subsection{Markov Decision Processes and Reinforcement Learning}

A finite-horizon Markov Decision Process (MDP) $\mdp$ is a model for sequential decision making over a number of time steps $t \in T$ defined by a tuple $(\states, \actions, P, D_0, \phi, \gamma, R)$ \cite{sutton-barto-rl}. $\states$ is a finite set of discrete states; $\{\actions_s\} \subseteq \actions$ is a set of actions available at state $s$; $P: \states \times \actions \times \states \rightarrow [0,1] $ is a model of the environment given as transition probabilities where $P(s_{t+1}|s_t,a_t)$ is the probability of transitioning to state $s_{t+1}$ from state $s_t$ after taking action $a_t \in \{\actions_{s_t}\}$ at time $t$. $D_0: \states \rightarrow [0,1]$ is a distribution over start states;  $\phi : \states \times \actions \times \states \rightarrow \reals^k$ is a mapping from the transitions to a $k$-dimensional space of features; $\gamma \in [0,1)$ is a discount factor; and $R: \states \times  \actions \times \states \rightarrow \reals$ is a scalar reward received by the agent for being in one state and transitioning to another state at time $t$, written as $R(s_t,a_t,s_{t+1})$. 

An agent acts within the environment defined by the MDP, generating a sequence of actions called a \emph{trajectory} of length $t$. Let $\tau = ((s_1,a_1, s_{2}),...,(s_{t-1},a_{t-1}, s_{t})) \in (\states \times \actions \times \states)^t$. We evaluate the quality of a particular trajectory in terms of the amount of reward accrued over the trajectory, subject to discounting. Formally, $R(\tau) = \sum^t_{i=1} \gamma^i R(s_i, a_i, s_{i+1})$. A policy, $\pi: \states \rightarrow \mathcal{P}(\mathcal{A})$ is a map of probability distribution to actions for every state such that $\pi(s,a)$ is the probability of taking action $a$ in state $s$. We can also write the probability of a trajectory $\tau$ under a policy as $\pi(\tau)$. The feature vector associated with trajectory $\tau$ is defined as the summation over all transition feature vectors in $\tau$, $\phi(\tau) = \sum_{(s_t, a_t, s_{t+1}) \in \tau} \phi(s_t, a_t, s_{t+1})$

The goal within an MDP is to find a policy $\pi^*$ that maximizes the expected reward, $J(\pi) = \mathbb{E}_{\tau \sim \pi}[R(\tau)]$ \cite{anwar-2021-icrl}. In the MDP literature, classical tabular methods are used to find $\pi^*$ including value iteration (VI). Such method
finds an optimal policy by estimating the expected reward for taking an action $a$ in a given state $s$, i.e., the $Q$-value of pair $(s,a)$, written $q(s,a)$. \cite{sutton-barto-rl}.

\subsection{Constrained MDPs and Inverse Reinforcement Learning}

We are interested in learning constraints from demonstrations. Our goal is to create agents that are able to be trained to follow constraints that are not explicitly prohibited in the MDP, but should be avoided \cite{FoMa19a}. \cite{scobee-2020-maximumlikelihood} discusses the importance of such constraints: an MDP $\mdp$ may encode everything necessary about driving a car, e.g. the dynamics of steering and movements, but often one wants to add additional general constraints such as \emph{avoid obstacles on the way to the goal}. These constraints are often non-Markovian and engineering a reward function that encodes these constraints may be a difficulty or impossible task \cite{vazquez-2018-taskfromdemonstrations}.

One approach for learning constraints from demonstrations is to use techniques from inverse reinforcement learning (IRL): given a set of demonstrated trajectories $\mathcal{D}$ of an agent in an environment $\mdp$ with an unknown reward function $\mathcal{M} \setminus R$, IRL provides a set of techniques for learning a reward function $\hat{R}$ that explains the agent's demonstrated behavior \cite{abbeel2004apprenticeship,Ng2000}. However, this technique has many drawbacks: often there are many reward functions that lead to the same behavior \cite{scobee-2020-maximumlikelihood}, the reward functions may not be interpretable \cite{vazquez-2018-taskfromdemonstrations}, and there may be issues such as reward hacking -- wherein the agent learns to behave in ways that create reward but are not intended by the designer -- an important topic in the field of AI safety \cite{amodei2016concrete,ray2019benchmarking} and value alignment \cite{FoMa19a,russell2015research}. 

We follow the framework of \citet{altman-1999-constrained} and \citet{anwar-2021-icrl} and define a Constrained MDP $\learnedmdp$ which is a nominal MDP $\nominalmdp$ with an additional cost function $\constraints: \states \times \actions \times \states \rightarrow \reals$ and a budget $\alpha \geq 0$. We can then define the cost of a trajectory to be $c(\tau) = \sum^t_{i=1} c(s_i, a_i, s_{i+1})$. Setting $\alpha = 0$ is enforcing \emph{hard constraints}, i.e., we must never trigger constrained transitions. In this work, unlike the work of both \citet{scobee-2020-maximumlikelihood} and \citet{anwar-2021-icrl}, we are interested in learning \emph{soft constraints} \cite{rossi2006handbook}. Under a soft constraints paradigm, each constraint comes with a real-valued penalty/cost and the goal is to minimize the sum of penalties incurred by the agent. 

Following \citet{scobee-2020-maximumlikelihood}, the task of constraint inference in IRL is defined as follows. Given a nominal MDP $\nominalmdp$ and a set of demonstrations $\mathcal{D}$ in ground-truth constrained world $\trueconstrainedmdp$, we wish to find the most likely set of constraints $\constraints$ that could modify $\nominalmdp$ to explain the demonstrations. We are concerned with three types of constraints:
\begin{description}
    \item [Action Constraints.] We may not want an agent to ever perform some (set of) action $a_i$.
    \item [Occupancy Constraints.] We may not want an agent to occupy a (set of) states $s_i$.
    \item [Feature Constraints.] Given a feature mapping of transitions $\phi$, we may not want an agent to perform an (set of) action in presence of specific state features.
\end{description}

Without loss of generality, we add the state and actions to the features. Hence, action and occupancy become specific cases of feature constraints. Note that the set of constraints is defined as a cost function over the set of transitions $\constraints: \states \times \actions \times \states \rightarrow \reals$. In \citet{scobee-2020-maximumlikelihood}, this definition is limited to a set of state-actions $\states \times \actions$ as they are assuming a deterministic setting and hence are able to define $\learnedmdp$ by substituting $\mathcal{A} = \{A_s\}$ with $\mathcal{A}^\mathcal{C} = \{A^\mathcal{C}_s\}$ in $\nominalmdp$. Finally, both \citet{scobee-2020-maximumlikelihood} and \citet{anwar-2021-icrl} propose a greedy approach to infer a set of constraint $\constraints$ that explains the demonstrations $D$ on $\trueconstrainedmdp$. In both \citet{scobee-2020-maximumlikelihood} and \citet{anwar-2021-icrl} the domain is restricted to deterministic MDPs, which we strictly generalize in this work; additionally \citet{scobee-2020-maximumlikelihood}, like our model, only works with discrete actions, while \citet{anwar-2021-icrl} works for both discrete and continuous action sets. We also  generalize to the non-deterministic setting; we additionally generalize to the setting of soft constraints, hence our task is to learn the cost function $\constraints$.

To test our methods, we use the same grid world setup as \citet{scobee-2020-maximumlikelihood}. Within our grid world example, shown in Figure \ref{fig:example_grid}, we have an action penalty of $-4$ for the cardinal directions, $-4 \times \sqrt{2}$ for taking the diagonal actions, and reaching the goal state has a reward of $10$.
In Figure \ref{fig:example_grid} we set the constraint costs to various values but in all our experiments we fix the constraint costs on the generated grids for states, actions, and features to be $-50$. Throughout we assume a non-deterministic world with a $10\%$ chance of action failure, resulting in a random action.

\subsection{Multi-Alternative Decision Field Theory}
\label{sec:mdft-back}

\begin{figure}[h]
\centering
\includegraphics[width=0.45\textwidth]{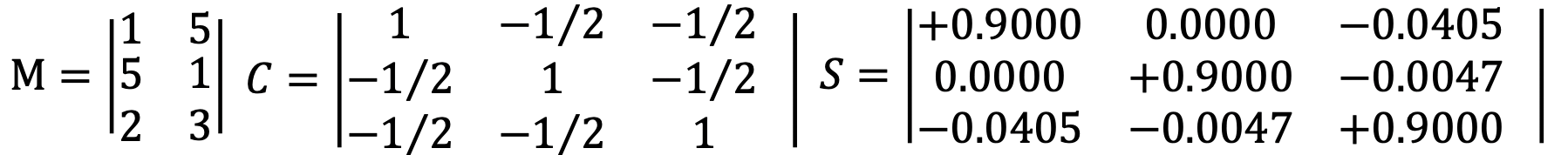}
\caption{Evaluation (M), Contrast (C), and Feedback (S) matrix.}
\label{fig:matrix}
\end{figure}

Multi-alternative Decision Field Theory (MDFT) is a dynamic cognitive approach that models human decision making based on psychological principles \cite{busemeyer2002survey,roe2001multialternative}. MDFT models preferential choice as an iterative cumulative process in which at each time instant the decision maker attends to a specific attribute to derive comparisons among options and update their preferences accordingly. Ultimately the accumulation of those preferences informs the decision maker's choice.  In MDFT an agent is confronted with multiple options and equipped with an initial personal evaluation for them along different criteria, called attributes. For example, a student who needs to choose a main course among those offered by the cafeteria will have in mind an initial evaluation of the options in terms of how tasty and healthy they look. More formally, an MDFT model is composed of the following \cite{roe2001multialternative}:

\textbf{Personal Evaluation}: 
Given set of options $O=\{o_1, \dots, o_k\}$ and set of attributes
$A=\{A_1, \dots, A_J\}$,
the subjective value of option $o_i$ on 
attribute $A_j$ is denoted by $m_{ij}$ and stored in matrix $\textbf{M}$. 
In our example, let us assume that the cafeteria options  are 
{\em Salad (S)}, {\em Burrito (B)} and {\em Vegetable pasta (V)}.
Matrix $\mathbf{M}$, containing the student's preferences,
could be defined as shown in Figure \ref{fig:matrix} (left),
where rows correspond to the options $(S,B,V)$ and the columns 
to the attributes $Taste$ and $Health$.

\textbf{Attention Weights}: Attention weights are used to express the attention allocated to each attribute at a particular time $t$ during the deliberation. We denote them by vector $\textbf{W}(t)$ where $W_j(t)$ represents the attention to attribute $j$ at time $t$. We adopt the common simplifying assumption that, at each point in time, the decision maker attends to only one attribute \cite{roe2001multialternative}. Thus, $W_j(t)\in\{0,1\}$ and $\sum_jW_j(t)=1$, $\forall t,j$. In our example, we have two attributes, so at any point in time $t$ we will have $\textbf{W}(t)=[1,0]$, or $\textbf{W}(t)=[0,1]$, representing that the student is attending to, respectively, $Taste$ or $Health$. The attention weights change across time according to a stationary stochastic process with probability distribution $\textbf{w}$, where $w_j$ is the probability of attending to attribute $A_j$. In our example, defining $w_1=0.55$ and $w_2=0.45$ would mean that at each point in time, the student will be attending $Taste$ with probability $0.55$ and $Health$ with probability $0.45$. In other words, $Taste$ matters slightly more than $Health$.

\textbf{Contrast Matrix}: Contrast matrix $\textbf{C}$ is used to 
compute the advantage (or disadvantage) of  an option with respect to the other options.
In the MDFT literature \cite{busemeyer1993decision,roe2001multialternative,hotaling2010theoretical}, $\textbf{C}$
is defined by contrasting the initial evaluation of one alternative against the 
average of the evaluations of the others, as shown for the case with three options in Figure \ref{fig:matrix} (center).


At any moment in time, each alternative in the 
choice set is associated with a {\bf valence} value. The valence for option $o_i$ 
at time $t$, denoted $v_i(t)$, represents its momentary advantage (or 
disadvantage) when compared with other options on some attribute 
under consideration. The valence vector for $k$ options $o_1, \dots, o_k$ at 
time $t$, denoted by column vector $\mathbf{V}(t) = [v_1(t), \dots, v_k(t)]^T$, is formed by $\textbf{V}(t) = \textbf{C}\times \textbf{M} \times \textbf{W}(t)$.
%
%
In our example, the valence vector at any time point in which $\textbf{W}(t)=[1,0]$, is 
$\textbf{V}(t) = [1-\nicefrac{7}{2}, 5-\nicefrac{3}{2}, 2-\nicefrac{6}{2}]^T$.

In MDFT, preferences for each option are accumulated across the iterations of the deliberation process until a decision is made. This is done by using \textbf{Feedback Matrix} $\mathbf{S}$, which defines how the accumulated preferences affect the preferences computed at the next iteration. This interaction depends on how similar the options are in terms of their initial evaluation expressed in $\mathbf{M}$. Intuitively, the new preference of an option is affected positively and strongly by the preference it had accumulated so far, while it is inhibited by the preference of similar options. This 
lateral inhibition decreases as the dissimilarity between options increases. Figure \ref{fig:matrix} (right) shows 
\textbf{S} for our example following the MDFT method in \cite{hotaling2010theoretical}. 
At any moment in time, the  preference of each alternative is calculated by 
$\textbf{P}(t + 1) = \textbf{S} \times \textbf{P}(t) + \textbf{V}(t + 1)$
where  $\textbf{S} \times \textbf{P}(t)$ is the contribution of the past preferences and 
$\textbf{V}(t + 1)$ is the valence computed at that iteration. Starting with $\textbf{P}(0)=0$,
preferences are then accumulated for either a fixed number of iterations (and the option with the highest preference is selected) or until the preference of an option reaches a given threshold. 
In the first case, MDFT models decision making with a \textit{specified} deliberation time, while, in the latter, it models cases where deliberation time is \textit{unspecified} and choice is dictated by the accumulated preference magnitude.
In general, different runs of the same MDFT model 
may return different choices due to the attention weights' distribution. 
In this way MDFT induces choice distributions over set of options and is capable of capturing well know behavioral effects such as the compromise, similarity, and attraction effects that have been observed in humans and that violate rationality principles \cite{busemeyer1993decision}.   

 






\begin{figure*}[ht]
\begin{minipage}[t]{.35\textwidth}
  \centering
  \includegraphics[width=\linewidth]{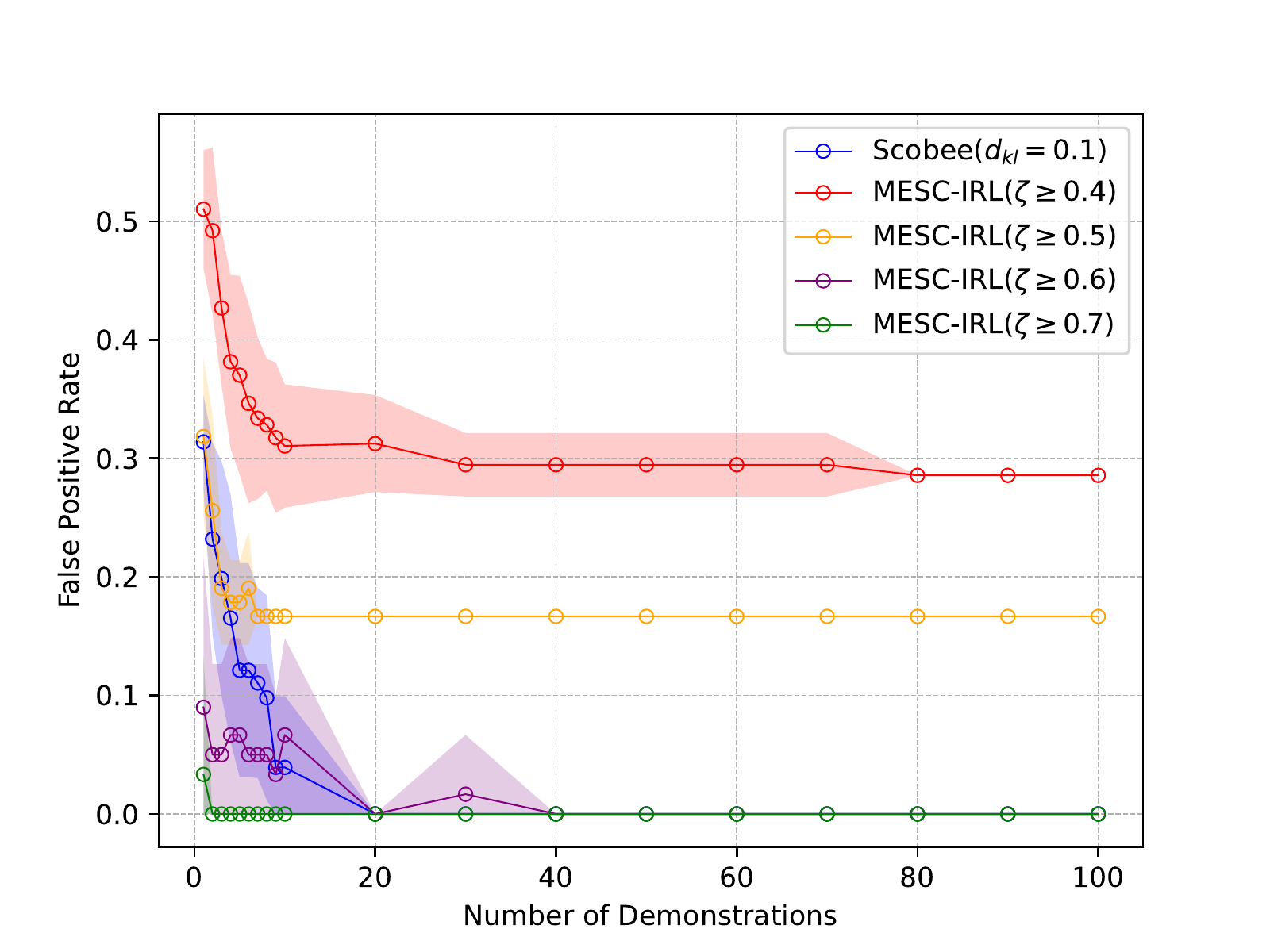}
\end{minipage}%
\hfill
\begin{minipage}[t]{.35\textwidth}
  \centering
  \includegraphics[width=\linewidth]{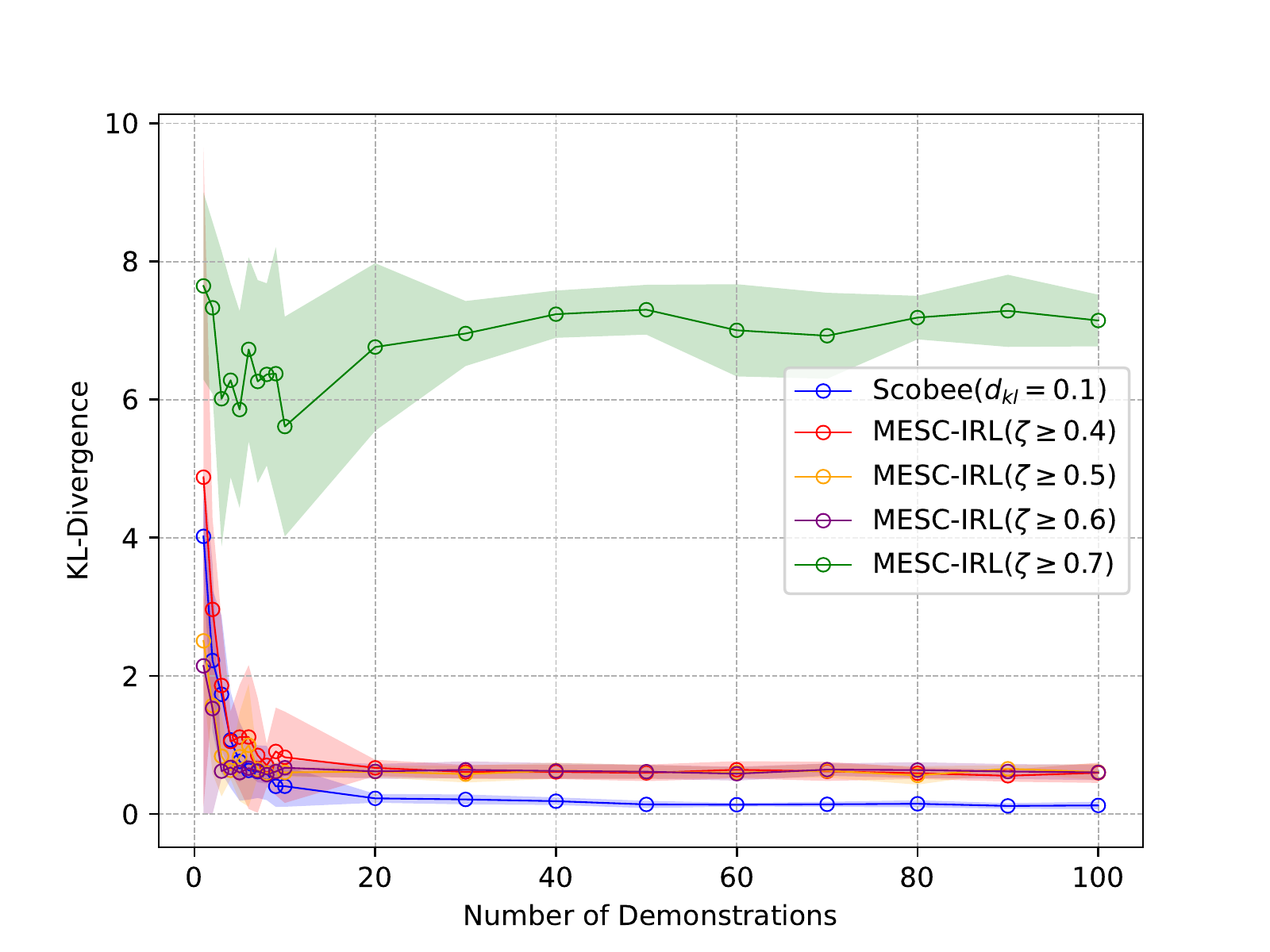}
\end{minipage}
\caption{Comparision of MESC-IRL to the best performing method of \citet{scobee-2020-maximumlikelihood} at recovering hard constraints in a deterministic setting according to false positive rate (left) and KL-Divergence from the demonstrations $\demonstrations$ (right) as we vary the number of demonstrations. Each point is the mean of 10 independent draws. }
\label{fig:scobee}
\end{figure*}

\section{Learning Soft Constraints From Demonstrations}
\label{sec:mesc-irl}

We now describe our method for learning a set soft constraints from a set of demonstrations $\demonstrations$ and a nominal MDP $\nominalmdp$. This is the first step in our goal of providing a flexible and expressive method for orchestrating trade-offs. The method described here generalizes the work of both \citet{scobee-2020-maximumlikelihood} and \citet{anwar-2021-icrl} to the setting of non-deterministic MDPs and soft constraints.

\subsection{MESC-IRL: Max Entropy Inverse Soft-Constraint Reinforcement Learning}

Following \citet{ziebart-2008-maxentropy}, our goal is to optimize a function that linearly maps the features of each transition to the reward associated with that transition, $R(s_t, a_t, s_{t+1}) = \omega  \phi(s_t, a_t, s_{t+1})$, where $\omega$ is the reward weight vector. \citet{ziebart-2008-maxentropy} propose a maximum entropy model for finding a unique solution ($\omega$) for this problem. Based on this model, the probability of finite-length trajectory $\tau$ being executed by an agent traversing an MDP $\mathcal{M}$ is exponentially proportional to the reward earned by that trajectory and can be approximated by:
\begin{equation*}
    \resizebox{0.90\linewidth}{!}{
    $P(\tau|\omega) \approx \frac{e^{\omega^T\phi(\tau)}}{Z(\omega)} \prod_{(s_t, a_t, s_{t+1}) \in \tau} P(s_{t+1} | s_t, a_t)$.
    }
\end{equation*}
The optimal solution is obtained by finding the maximum likelihood of the demonstrations $\demonstrations$ using this probability distribution: $\omega^* = \underset{\omega}{\text{argmax}} \sum_{\tau \in \mathcal{D}} \log P(\tau|\omega)$.

We extend the problem defined in \citet{scobee-2020-maximumlikelihood} to that of learning a set of \emph{soft} constraints which best explain a set of observed demonstrations. This allows us to move from the notion of a constraint forbidding an action or a state to that of a soft constraint imposing a penalty proportional to the gravity of its violation. In other words, given access to $\nominalmdp$ and a set of demonstrations $\demonstrations$ in ground-truth constrained MDP  $\trueconstrainedmdp$ we want to find the costs $\constraints$. More formally, we define the residual reward function $R^{\mathcal{R}}: \mathcal{S} \times \mathcal{A} \times \mathcal{S} \rightarrow \mathbb{R_+}$ as a mapping from the transitions to the penalties. We can now formally define our soft-constrained MDP $\learnedmdp$ as follows:
\begin{definition}
Given $\nominalmdp = \langle \mathcal{S}, \mathcal{A}, P, \mu, \phi, R^\mathcal{N} \rangle$ we define  soft-constrained MDP $\learnedmdp = \langle \mathcal{S}, \mathcal{A}, P, \mu, \phi, R^\mathcal{C} \rangle$ where $R^\mathcal{C} = R^{\mathcal{N}} - R^{\mathcal{R}}$.  
\end{definition}
\noindent
Thus, the goal of our task is to find a residual reward function $R^{\mathcal{R}}$ that maximizes the likelihood of the demonstrations $\mathcal{D}$ given the nominal MDP $\mathcal{M}^\mathcal{N}$.

Our solution is based on adapting Maximum Causal Entropy Inverse Reinforcement learning \cite{ziebart-2008-maxentropy, ziebart-2010-causalentropy} to soft-constrained MDPs. 
Following the setting of  \citet{ziebart-2008-maxentropy}  we can write the reward function $R^\mathcal{N}$ (resp. $R^\mathcal{C}$) of $\mathcal{M}^\mathcal{N}$ (resp. $\learnedmdp$) as a linear combination of the transitions:
    $R^\mathcal{N}(s_t, a_t, s_{t+1}) = \omega^\mathcal{N} \phi(s_t, a_t, s_{t+1})$ and 
    $R^\mathcal{C}(s_t, a_t, s_{t+1}) = \omega^\mathcal{C}  \phi(s_t, a_t, s_{t+1})$.
%
As, both reward functions $R^\mathcal{N}$ and $R^\mathcal{C}$ are linear, $R^\mathcal{R}$ should be linear as well
    $R^\mathcal{R} = \omega^\mathcal{R}\phi(s_t, a_t, s_{t+1})$.
%
From this formulation of $R^\mathcal{R}$ we can infer that the reward vectors follow 
    $\omega^\mathcal{C} = \omega^\mathcal{N} - \omega^\mathcal{R}$.

At this point we can use Max Entropy IRL for learning a reward function compatible with the trajectories in $\mathcal{D}$. The gradient for maximizing the likelihood in this setting is defined as in \citet{ziebart-2008-maxentropy}:
\begin{equation}\label{const_gradient}
    \resizebox{0.90\linewidth}{!}{
    $\nabla_{\omega^\mathcal{C}} \mathcal{L}(\mathcal{D}) = \mathbb{E}_{\mathcal{D}}[\phi(\tau)] - \sum_{(s_t, a_t, s_{t+1})} D_{s_t, a_t, s_{t+1}}  \phi(s_t, a_t, s_{t+1})$
    }
\end{equation}
Where $D_{s_t, a_t, s_{t+1}}$ is the expected feature frequencies for transition $(s_t, a_t, s_{t+1})$ using the current $\omega^\mathcal{C}$ weights. Given that the reward vectors follow $\omega^\mathcal{C} = \omega^\mathcal{N} - \omega^\mathcal{R}$, we can write 
    $\nabla_{\omega^\mathcal{C}} = - \nabla_{\omega^\mathcal{R}}$.
%
Finally, by substituting this in Eq. \ref{const_gradient} we obtain the gradient of likelihood of the constrained trajectories with respect to $\omega^\mathcal{R}$: 
\begin{equation}
    \resizebox{0.90\linewidth}{!}{
    $\nabla_{\omega^\mathcal{R}} \mathcal{L}(\mathcal{D}) = \sum_{(s_t, a_t, s_{t+1})} D_{s_t, a_t, s_{t+1}}  \phi(s_t, a_t, s_{t+1}) - \mathbb{E}_{\mathcal{D}}[\phi(\tau)]$
    }
\end{equation}
As we estimate the residual rewards with respect to the nominal rewards, these rewards are automatically scaled to be compatible with the nominal rewards.

\subsection{Generalizing From Penalties to Probabilities}\label{sec:prob}

The estimated penalties from the previous section can effectively guide an agent to navigate the environment optimally as well as provide estimates of the cost of the constraints scaled to the value of the original reward signal. However, there may be instances, such as when comparing with hard constraints, where we desire \emph{probabilities} that a particular action is constrained. Having probabilities allows us to compare constraints across environments with possibly different scales, allows us to use this information to guide our policies, and allows us to evaluate the confidence we have in a particular constraint. In this section we describe a method to transition from penalties to probabilities, as well as a generalized method to extract these probabilities based on a subset of the features of the environment, which can facilitate transfer learning between domains.

Intuitively, a transition where the residual reward, i.e., the penalty, is significantly larger than zero is more likely to be a constraint. We estimate the significance of a penalty by scaling it to the standard deviation of the mean learned reward. Therefore, we assume that a transition penalty is a random variable, denoted by $\mathbb{C} \sim logistic(\sigma_{pooled}, \sigma_{pooled})$, following a logistic distribution with standard deviation $\sigma_{pooled}$, where $\sigma_{pooled} = \sqrt{\nicefrac{(\sigma_{\mathcal{N}}^2 + \sigma_{\mathcal{C}}^2)}{2}}$ and $\sigma_{\mathcal{N}}$ and $\sigma_{\mathcal{C}}$ are the standard deviations of the rewards in the nominal and learned constrained worlds, respectively. Informally, when penalties are close to zero, we want their probabilities to be small. To do this we set the mean of the distribution to be $\mu=\sigma_{pooled}$.

We now want to reason about a random variable $\zeta$ that indicates our belief that the transition $(s_t, a_t, s_{t+1})$ is forbidden. Hence using the above probability distribution we can define the probability of constraint given a transition as:
\begin{equation*}
    \resizebox{0.90\linewidth}{!}{
    $\zeta \equiv  P\left(\mathbb{C} \leq R^{R}(s_t, a_t, s_{t+1})\right) = sigmoid\left(\frac{R^{R}(s_t, a_t, s_{t+1}) - \sigma_{pooled}}{\sigma_{pooled}}\right)$
    }
\end{equation*}

In our formulation, the residual rewards only depend on the features associated with them. Hence, we can use this fact to reason about constraints over only a subset of features $\mathbf{f}$, e.g., only color or state position. Let $\phi_f \subseteq \phi$ be the subset of features we are concerned with. In our grids we represent $\phi$ with a vector of length 92. The first 81 elements represent the states, the next 8 represent the actions, and the last 3 represent the colors. So if we are interested in only learning about constraints over the colors, $\phi_f$ will be a vector equal to the last three elements of $\phi$ that is $\phi_{color} \equiv \phi_{90, 91, 92}$. 

Let $\phi_{\mathbf{f}}$ and $\omega^{R}_{\mathbf{f}}$ be the feature function and residual feature weight vector for $\mathbf{f}$. We can now define the probability of a feature value to be constrained 
as: 
\begin{equation*}
    \resizebox{0.8\linewidth}{!}{
    $\zeta_{\mathbf{f}} \equiv P(\mathbb{C} \leq \omega^{R}_{\mathbf{f}} \phi_{\mathbf{f}}) = sigmoid\left(\frac{\omega_{\mathbf{f}} \phi_{\mathbf{f}} - std_{pooled}}{std_{pooled}}\right)$.
    }
\end{equation*}


\subsection{Experimental Evaluation of MESC-IRL}

In this section we empirically validate our method for soft constraint learning against both the method of \citet{scobee-2020-maximumlikelihood} for learning hard constraints in deterministic settings as well as on learning soft constraints in non-deterministic settings. Figure \ref{fig:scobee} shows the performance of MESC-IRL compared to the method proposed by \citet{scobee-2020-maximumlikelihood} on the same metrics from their paper: false positives, i.e., predicting a constraint when one does not exist, and KL-Divergence from the demonstrations set $\demonstrations$. For this test we use the same single grid, hard constraints, and a deterministic setting to allow for a direct comparison. We generate 10 independent sets of 100 demonstrations and report the mean. In order to decide if the values returned by MESC-IRL represent a hard constraint, we threshold the value of $\zeta$ at various levels and plot the comparison to the best result from \citet{scobee-2020-maximumlikelihood}. MESC-IRL with $\zeta \geq 0.6$ performs better than existing methods when the number of demonstrations is low, about the same when there are more demonstrations, and is able to also work for soft constraints and non-deterministic settings.


In order to evaluate MESC-IRL on soft constraints we need to adapt the notion of false positives and false negatives. Let a false positive $fp$ be:
\begin{equation*}
    \resizebox{0.60\linewidth}{!}{
    $fp = \frac{\Big|\big\{x \in \constraints \mid c(x) = 0 \wedge (\zeta_{\mathcal{C}}(x) - \zeta_{\mathcal{C}^*}(x) > \chi)\big\} \Big|}{\text{Num. Constraints}}$.
    }
\end{equation*}
Where $\zeta_{\mathcal{C}}(x)$ and $\zeta_{\mathcal{C}^*}(x)$ are the predicted and true probability of transition $x$ being constrained as described in Section \ref{sec:prob}, and $\chi$ is a value in [0,1]. Intuitively, we count a constraint as a false positive whenever there is no constraint in $\trueconstrainedmdp$ and the predicted probability exceeds the true probability by more than the threshold $\chi$. We can adapt the notion of false negatives, $fn$ in the same way by taking $c(x) \neq 0$.

\begin{figure*}[t]
\begin{minipage}[t]{.33\textwidth}
  \centering
  \includegraphics[width=\linewidth]{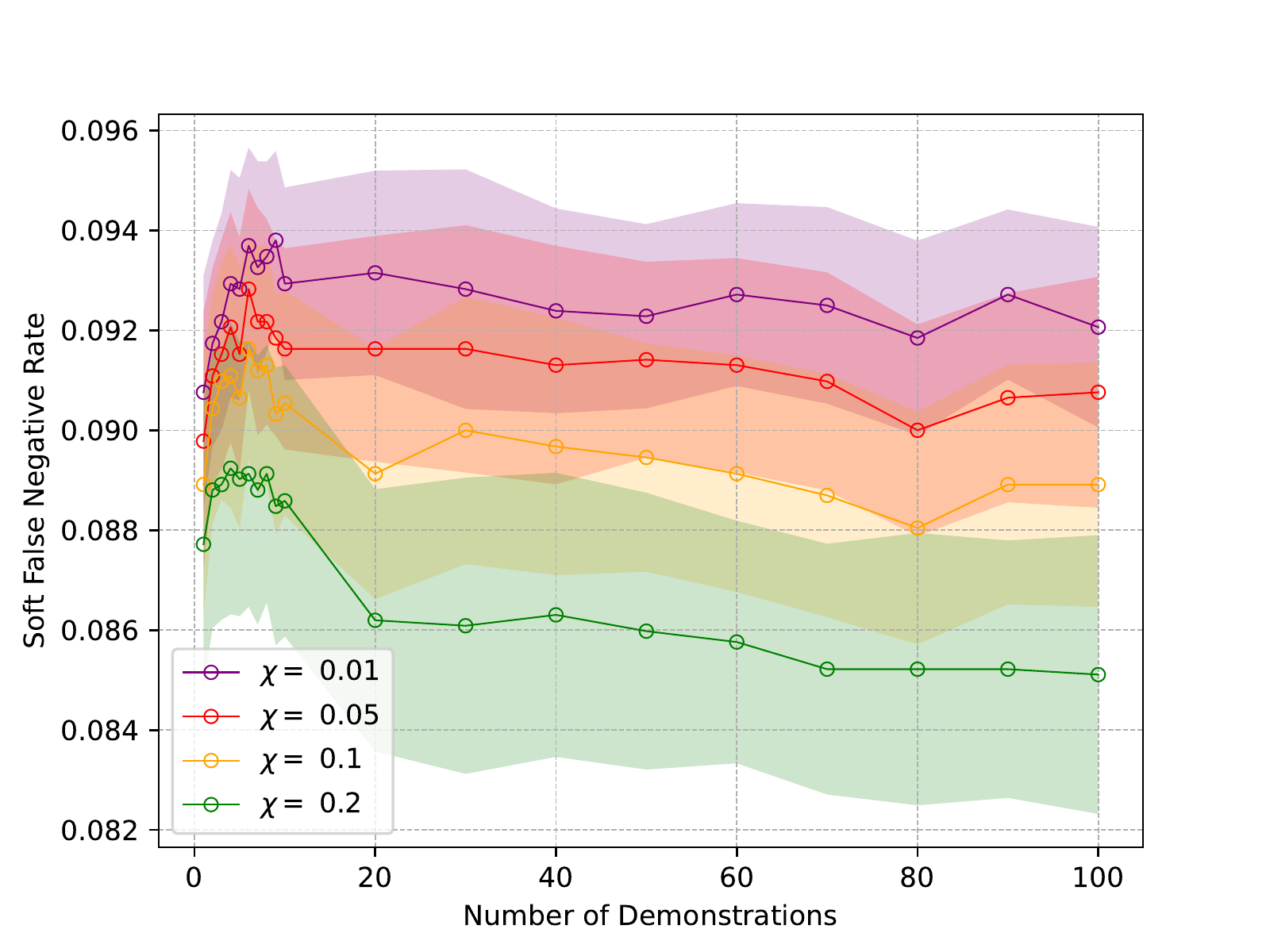}
\end{minipage}%
\hfill
\begin{minipage}[t]{.33\textwidth}
  \centering
  \includegraphics[width=\linewidth]{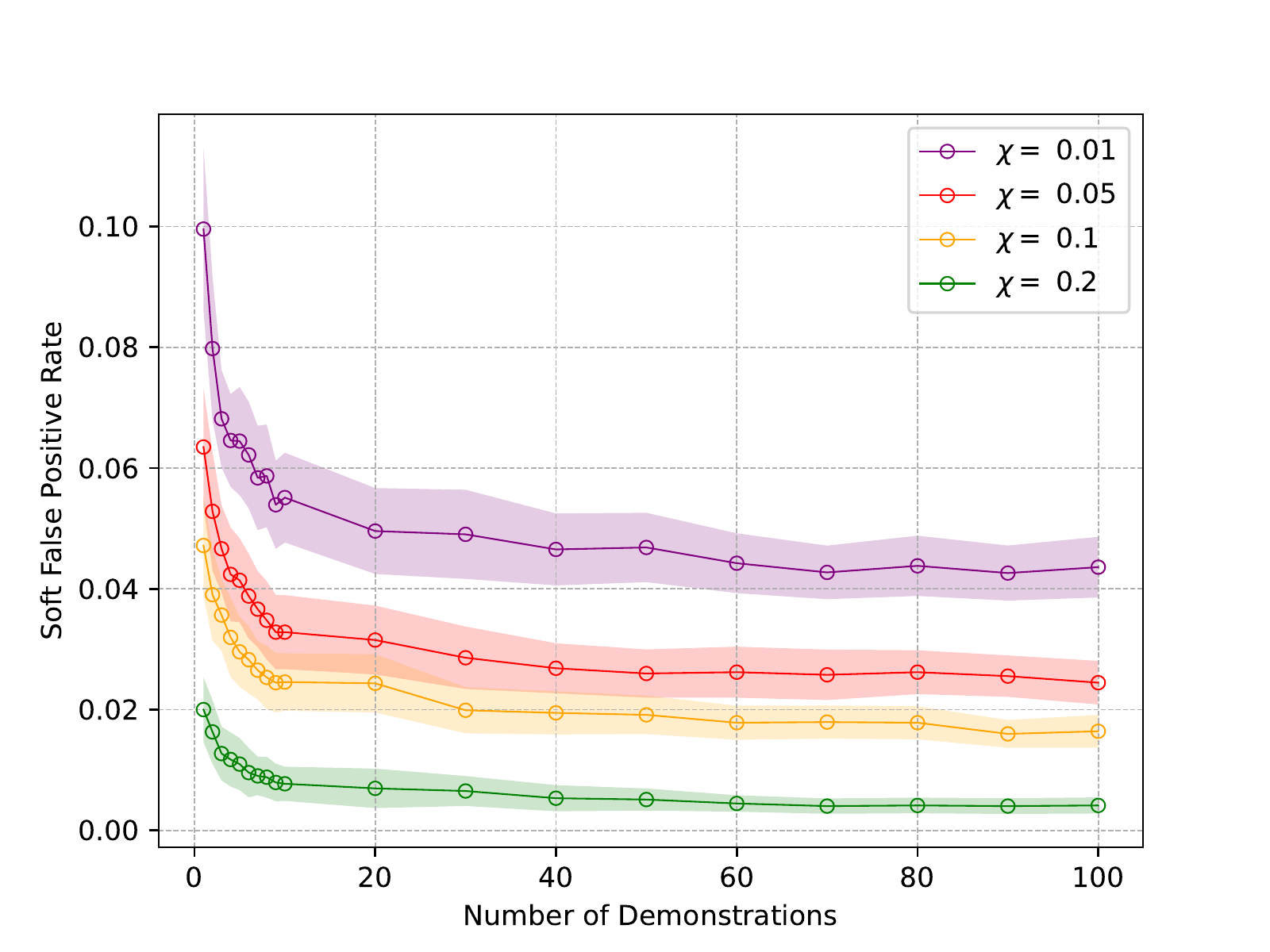}
\end{minipage}%
\hfill
\begin{minipage}[t]{.33\textwidth}
  \centering
  \includegraphics[width=\linewidth]{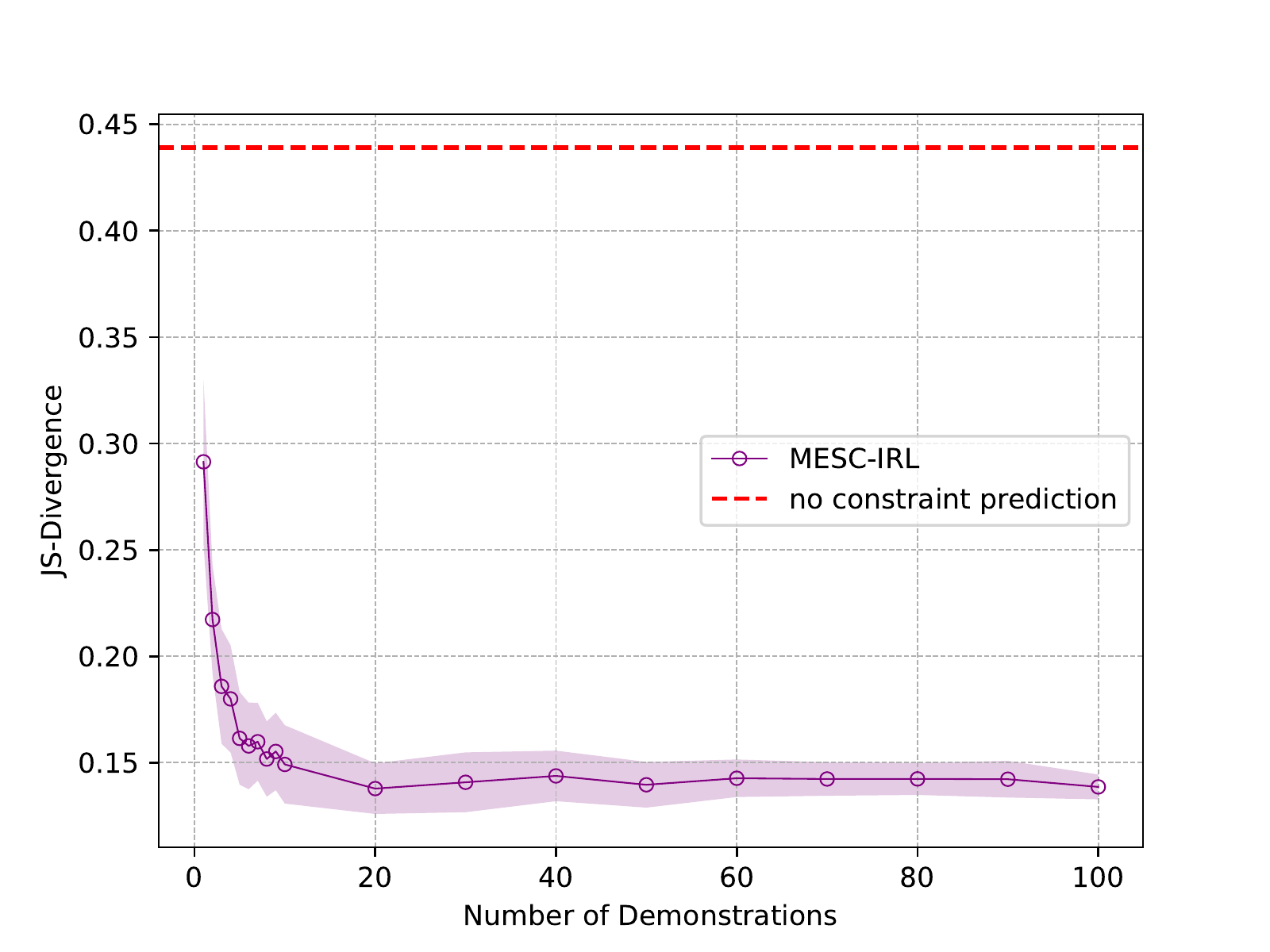}
\end{minipage}
\caption{Performance of MESC-IRL on recovering soft constraints in non-deterministic settings according to false negatives (left), false positives (center), and JS-Divergence to $\demonstrations$ (right). We see that across all these settings we are able to accurately recover constraints and generate behavior similar to the $\demonstrations$ even with few demonstrations.}
\label{fig:soft_performance}
\end{figure*}

Figure \ref{fig:soft_performance} shows the results of our tests on recovering soft constraints in non-deterministic settings with random grids, results for deterministic settings can be found in the Appendix. For these tests we choose a start and a goal state randomly at least 8 moves apart, set 6 states for blue, 6 for green randomly, and select 6 randomly constrained states; all penalties are set to $-50$. Again we take 10 sets of 100 demonstrations. We see a strong decrease in both false positives and false negatives as the number of demonstrations grows. We see that in general, and even more so  when the optimal threshold $\chi = 0.2$ is selected, our method almost never adds constraints that are not present in the ground truth and rarely underestimates the probability of existing ones, even for relatively small demonstration sets. Likewise our method is able to generate trajectories very close to $\demonstrations$, showing that we are able to recover both constraints and behavior even with soft constraints in non-deterministic setting. Hence MESC-IRL is able to work across a variety of settings and accurately capture demonstrated constraints.





\section{Orchestrating Goals and Constraints}

Often humans are confronted with decisions that require making trade-offs between collective norms and personal objectives \cite{FoMa19a,noothigattu-2019-ethicalvalues}. In this section we investigate different ways to model this orchestration in constrained grid environments. We consider different methods for combining policies $\pi_n$ for the nominal $\nominalmdp$ and $\pi_c$ for the learned constrained $\learnedmdp$.
%
%
For every state action pair $(s,a)$ we consider vectors $\langle sq_n(s,a_i)\rangle$ and $\langle sq_c(s,a_i)\rangle$ with $i\in\{1, \ldots, k\}$ where 
$sq_n(s,a_i)$ (resp. $sq_c(s,a_i)$) represents the probability of choosing action $a_i$ in state $s$ according to policy $\pi_n$ (resp. $\pi_c$). They are obtained by taking the softmax of the Q-values for each policy:



\begin{description}
    \item [Greedy ($\pi_{G}$):] Let $\pi_{G}(s) = a$, where each $a$ is the one with highest Q-value, $a$=$\underset{a \in \mathcal{A}_s}{argmax}$ $max\{q_c(s,a),q_n(s,a)\}$.
    
    \item [Weighted Average ($\pi_{WA}$):] Given weight vector $(w_n,w_c)$  with $w_n$,$w_c \in [0,1]$ and $w_c+w_n=1$, action $a=\pi_{WA}(s)$ is chosen according to probability distribution $p_{WA}(a_i)=w_n sq_n(s,a_i) + w_c sq_c(s,a_i)$.

    \item [MDFT ($\pi_{MDFT}$):] Action $a=\pi_{MDFT}(s)$ is chosen via an MDFT model where: \textbf{M} is a $k \times 2$ matrix where rows (i.e., options) correspond to actions and columns (i.e., attributes) correspond to $\nominalmdp$ and $\learnedmdp$. The $i$-th element of the respective world column is $sq_n(s,a_i)$ (resp., $sq_c(s,a_i)$), i.e., we are using the probability of choosing an action as a proxy of its preference. The weight vector $(w_n,w_c)$ is defined as for $\pi_{WA}$, and serves as probability distribution $\mathbf{w}$ defining how attention shifts between attributes during deliberation. Matrices $\mathbf{C}$ and $\mathbf{S}$ are defined in the standard way as described in Section \ref{sec:mdft-back}. When reaching state $s$, an MDFT deliberation process is launched to decide which action should be chosen. At each step the focus is shifted to $\nominalmdp$ or $\learnedmdp$ according to probability distribution $(w_n,w_c)$, and the preferences of the actions according to the selected attribute are accumulated as per Section \ref{sec:mdft-back}.
\end{description}  

Informally, Greedy is a deterministic approach that takes the most promising action, WA allows the agent to prioritize the pursuit of the goal state and satisfying constraints via a new policy obtained by considering the weighted average of the nominal and constrained distributions, and the MDFT-based orchestrator uses MDFT to chose at each step an action as suggested by the MDFT machinery.

\begin{figure*}[t]
\begin{minipage}[t]{.33\textwidth}
  \centering
  \includegraphics[width=\linewidth]{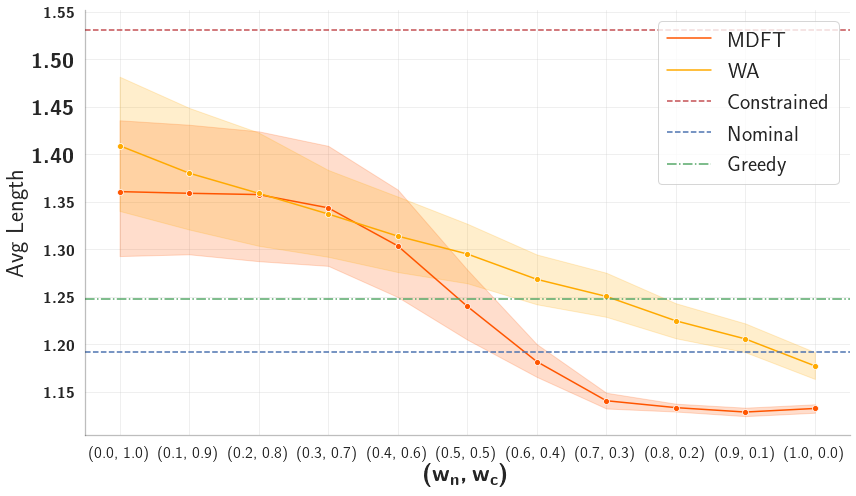}
\end{minipage}
\hfill
\begin{minipage}[t]{.33\textwidth}
  \centering
  \includegraphics[width=\linewidth]{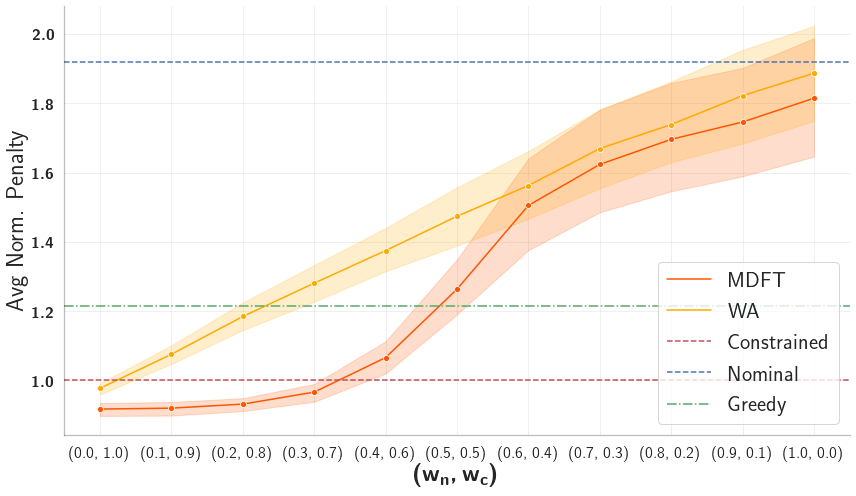}
\end{minipage}%
\hfill
\begin{minipage}[t]{.33\textwidth}
  \centering
  \includegraphics[width=\linewidth]{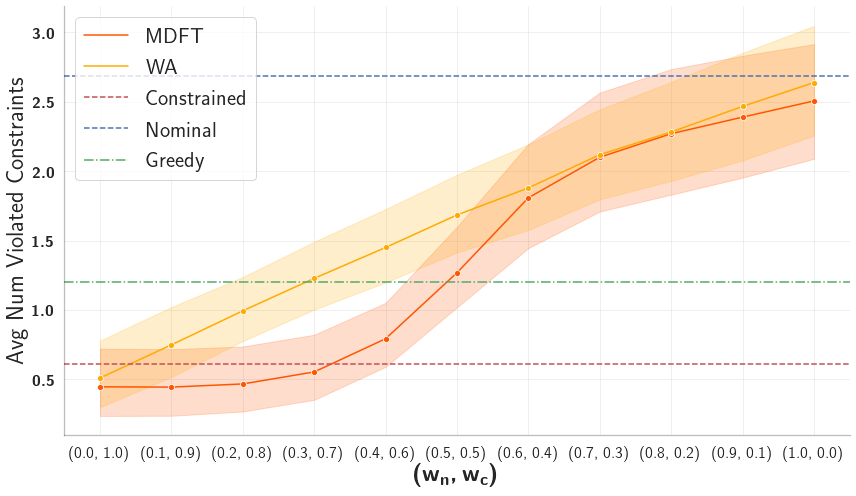}
  \label{fig:avg_vc}
\end{minipage}%

\caption{Comparison of the three orchestrators (Greedy, WA, and MDFT) on metrics including average path length (left), normalized penalty (lower is better, center), and average number of violated constraints (lower is better, right). We see that across all the metrics the MDFT orchestration is able to generate shorter paths that incur less penalty and violate fewer constraints.}
\label{fig:traj_performance}
\end{figure*}

\subsection{Comparison of Orchestration Methods}

We first compare theoretically the expressive power of the three orchestrators. We focus on a single state $s$ and consider how the policies  compare in terms of being able to model a given distribution over the actions available in $s$. We start by considering the Greedy orchestrator that is deterministic and will pick a fixed action $a$ in state $s$. Both WA and MDFT can model the Greedy policy by shifting all the weight to the environment where the maximum value is obtained and zeroing all preferences except for that of action $a$. A formal description is provided in the Appendix.
%
%
%
%
This observation, along with the fact that MDFT and WA are non-determistic, allows us to conclude that Greedy is strictly less expressive than the other two orchestrators.

Turning to the comparison between MDFT and WA, we can prove the following statement.

\begin{theorem}
Given any state $s$, there exist choice probability distributions over the actions available in $s$ that can be modeled by MDFT but not by WA.
\end{theorem}

We use an instance of the well known compromise effect \cite{busemeyer2002survey} according to which a compromising alternative tends to be chosen more often by humans than options with complementary preferences with respect  to the attributes. Consider the case of state $s$ with three actions $a_1$, $a_2$ and $a_3$. Let us assume that, for example, 
$sq_n(s,a_1)=\nicefrac{1}{6}$, $sq_n(s,a_2)=\nicefrac{1}{3}$ $sq_n(s,a_3)=\nicefrac{1}{2}$ and 
$sq_c(s,a_1)=\nicefrac{1}{2}$, $sq_n(s,a_2)=\nicefrac{1}{3}$ $sq_n(s,a_3)=\nicefrac{1}{6}$. 
According to the compromise effect humans will tend to choose $a_2$ more often than $a_1$ and $a_3$.
Such a choice distribution over the actions can be modeled by an MDFT defined over option set $\{a_1, a_2,a_3\}$, with two attributes and weights $w_n=0.55$ and $w_c=0.45$
\cite{busemeyer2002survey}. 
However, if we now consider WA, we can see that there is no way to define weights $(w_n,w_c)$ such that the corresponding weighted average probability satisfies
$w_n sq_n(s,a_2) + w_c sq_c(s,a_2)> max\{w_n sq_n(s,a_1) + w_c sq_c(s,a_1), w_n sq_n(s,a_3) + w_c sq_c(s,a_3)\}$. Thus, this distribution over actions cannot be modeled by the WA.

On the other hand, if we consider MDFTs in general, i.e. without the restriction of having two attributes, we can model any distribution. Intuitively this is achieved by defining an MDFT model over $k$ actions and with $k$ attributes where the weight of the $i$-th attribute corresponds to the  probability of the $i$-th action. Matrix $\mathbf{M}$ is set to the identity matrix and deliberation is halted after one iteration; see the Appendix for details.  

\begin{theorem}
\label{mdft-gen}
Given a $s$ and the set $\mathcal{A}_s$ of actions available in $s$,
consider a probability distribution $p$ defined over $\mathcal{A}_s$.
We can define an MDFT model where the set of options corresponds to $\mathcal{A}_s$ and the induced choice probability distribution coincides with $p$.
\end{theorem}

As a consequence, MDFT is general enough to express the probability distributions induced over the actions by WA.
Whether this is true also in the case of MDFT with only two attributes, as used in $\pi_{MDFT}$, remains an open theoretical question. However, we can see this experimentally:
in \citet{taher} the authors propose an RNN-based approach that starts from samples of a choice distribution and recovers parameters of an MDFT model, minimizing the divergence between the original and MDFT-induced choice distributions. We adapt their code\footnote{Available at \url{https://github.com/Rahgooy/MDFT}} and generate 100 
instances of WA distributions starting from random $sq_n$ and $sq_c$ distributions and $(w_n,w_c)$ weights. For each of these instances we generate 100 samples (i.e, chosen actions). We fix the  $sq_n$ and $sq_c$ values as parameters for the $\mathbf{M}$ matrix and learn the attention weight distribution $\mathbf{w}$ using 300 learning iterations. We use the learned MDFT model to generate a choice distribution over the actions with a stopping criteria of 25 deliberations steps. The observed average JS divergence between the original WA distributions and the ones induced by learned MDFT is 0.024 with standard error 0.0013; showing experimentally we can learn weights for an MDFT model to replicate any choice distribution of WA. 











\subsection{Experimentally Evaluating Orchestrators}

We compare the orchestrators empirically with the goal of testing if the combination of MESC-IRL with the orchestration techniques can be leveraged to create agents that trade-off between conflicting objectives like humans.

We start by generating 100 different non-deterministic nominal worlds, $\nominalmdp$, as described in Sections \ref{sec:prelim} and \ref{sec:mesc-irl}. We learn, via VI on $\trueconstrainedmdp$ the optimal policy in the (ground truth) constrained world, denoted $\optconstrained$ and, similarly to  \citet{scobee-2020-maximumlikelihood}, we use it  to generate sets of 200 demonstrations, $\demonstrations$. We then pass $\demonstrations$ to MESC-IRL, and the learned constraints are added to $\nominalmdp$ yielding the learned constrained MPD, \learnedmdp.
%
%
We use VI on \nominalmdp and \learnedmdp to obtain  $\pi_n$ and $\pi_c$, and  we consider different ways to prioritize them by sweeping the weight values $(w_n,w_c)$ from $(0,1)$ to $(1,0)$ in steps of $0.1$.
Note that at $(1,0)$ (resp. $(0,1)$) WA is equivalent to $\pi_n$ (resp. $\pi_c$), and that in both cases MDFT becomes deterministic, picking the action with highest Q-value.


For all our results we generated 200 trajectories for each step and method (including $\pi_n$ and $\pi_c$, denoted as Nominal and Constrained in Fig. \ref{fig:traj_performance}), and for each of the 100 random worlds. We first perform the Kolmogorov-Smirnov test to see if the trajectories generated by WA and MDFT induce the same distribution ($\text{H}_0$). We reject $\text{H}_0$ at every weight step with $p\_value \leq 0.01$, thus the two techniques induce statistically significantly different choice distributions.

\begin{figure}
    \centering
    \includegraphics[width=0.80\linewidth]{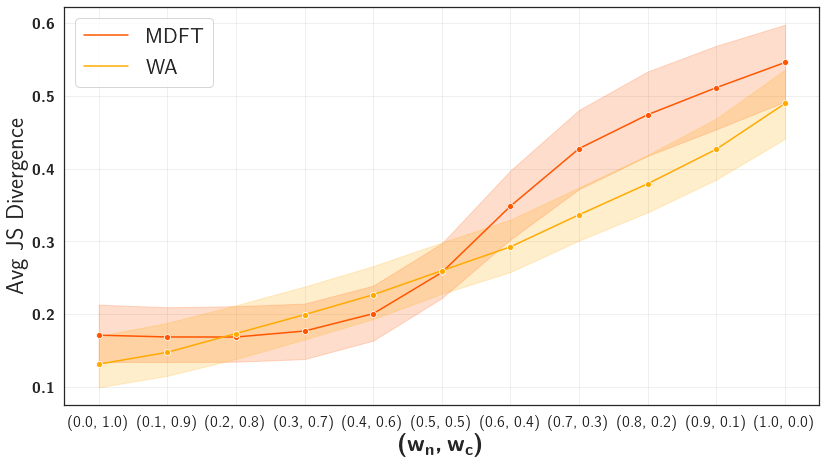}
    \caption{Average JS Divergence between policies generated with $\pi_c$ and orchestrators MDFT and WA.}
    \label{fig:avg_js}
\end{figure}




Figure \ref{fig:traj_performance} (left) shows the average length of trajectories produced by the orchestrators normalized so that $1.0$ is the shortest path between the start and goal state; (center) we scale the penalty by the average penalty for trajectories in \learnedmdp, lower is better; (right) we show the average number of violated constraints.
Across all these metrics, the MDFT agent is performing better than WA by always reaching the goal in a smaller number of steps no matter the configuration of the orchestrator. We can also see that the MDFT agent violates fewer constraints and accumulates lower penalties. 




Finally, in Figure \ref{fig:avg_js}  we show the JS Divergence between the trajectories generated by $\optconstrained$ and the trajectories generated by MDFT and WA, as the weight vector varies.  For both agents, the divergence is small on the left and grows moving to the right, as constraints become less important. This is not surprising, since the reference trajectories are generated using $\optconstrained$. Furthermore, we note that the MDFT advantage is more significant when $w_c$ is larger, that is when constraints matter more. An explanation for this is that a large value $w_c$ results in more MDFT deliberation steps to be focused (exclusively) on preferences relative to the constrained world. In WA, the averaging of the values underlying the policies, although weighted, is not able to maintain the importance of the constraints.







\section{Conclusions and Future Work}

We proposed a novel and general constraint learning method combined with a unique agent architecture aimed at learning constraints from demonstrations and exhibiting human-like trade-offs in environments with competing objectives. Our theoretical and experimental results show that  the MDFT-based method exhibits superior expressive power and performance both in terms of the quality of the produced trajectories as well as capability of capturing initial demonstrations. Another important features of this cognitive approach is its effectiveness in capturing behavioral traits of humans decision making, a key factor for real life applications involving human-generated demonstrations and decisions. Our MESC-IRL approach is a general approach for learning soft constraints over actions, states, and features in non-deterministic decision making environments. 

We plan to run experiments with human decision makers and employ also learning-based orchestrators \citet{noothigattu-2019-ethicalvalues}. We are also working on a novel multi-agent architecture with several orchestrators acting as either reactive or deliberate agents, e.g., the system 1 / system 2 model of  \citet{kahneman2011thinking} with a meta-cognitive agent to arbitrate, with the goal of further advancing the performance and generality of decision making agents.


\clearpage
\onecolumn
\appendix

\clearpage
\section*{Additional Material for \\ Making Human-Like Trade-offs in Constrained Environments by Learning from Demonstrations}

\bigskip
\section{Additional Graphs For Learning Constraints}

\begin{figure*}[h]
\begin{minipage}[t]{.48\textwidth}
  \centering
  \includegraphics[width=\linewidth]{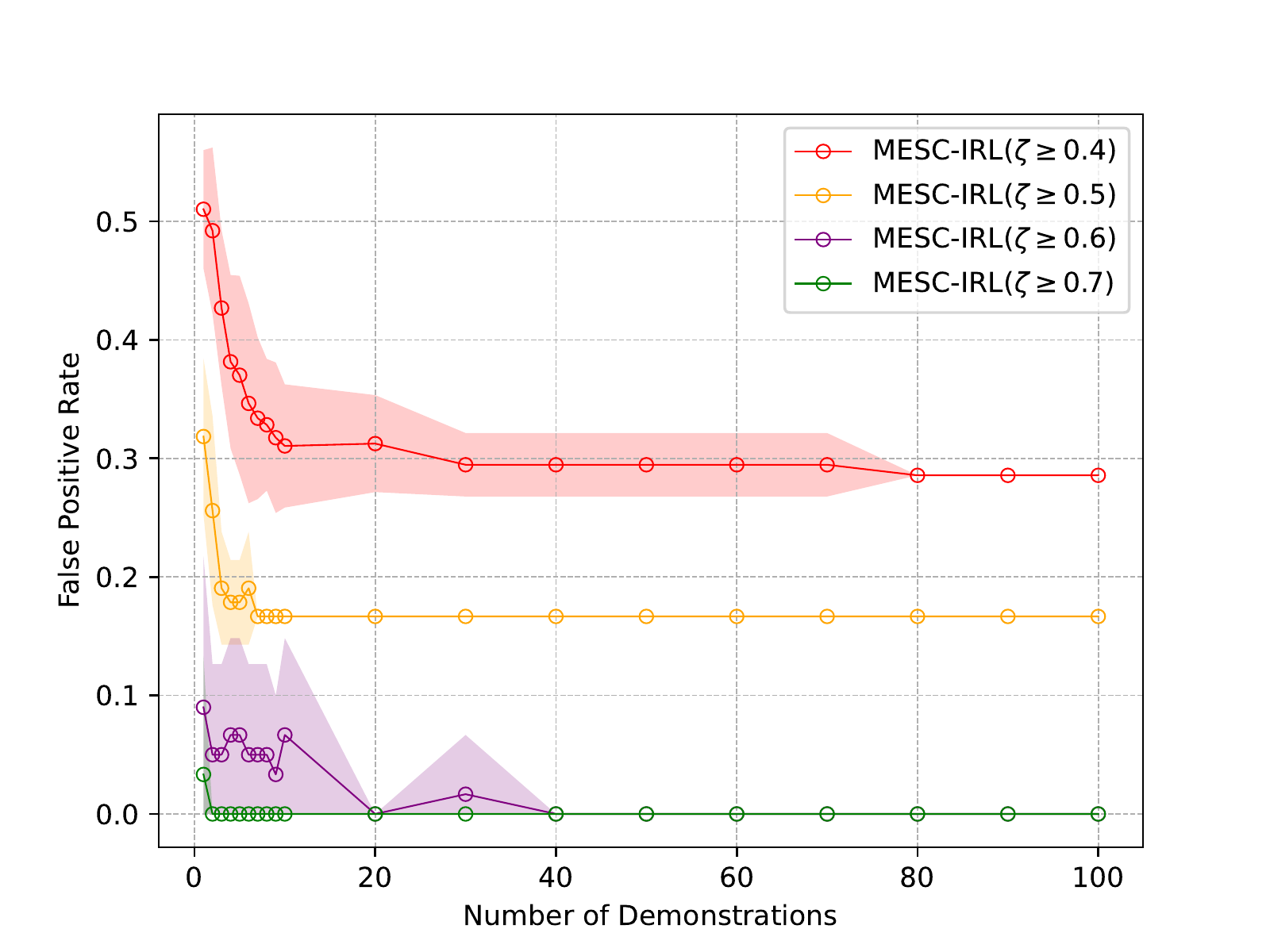}
\end{minipage}%
\hfill
\begin{minipage}[t]{.48\textwidth}
  \centering
  \includegraphics[width=\linewidth]{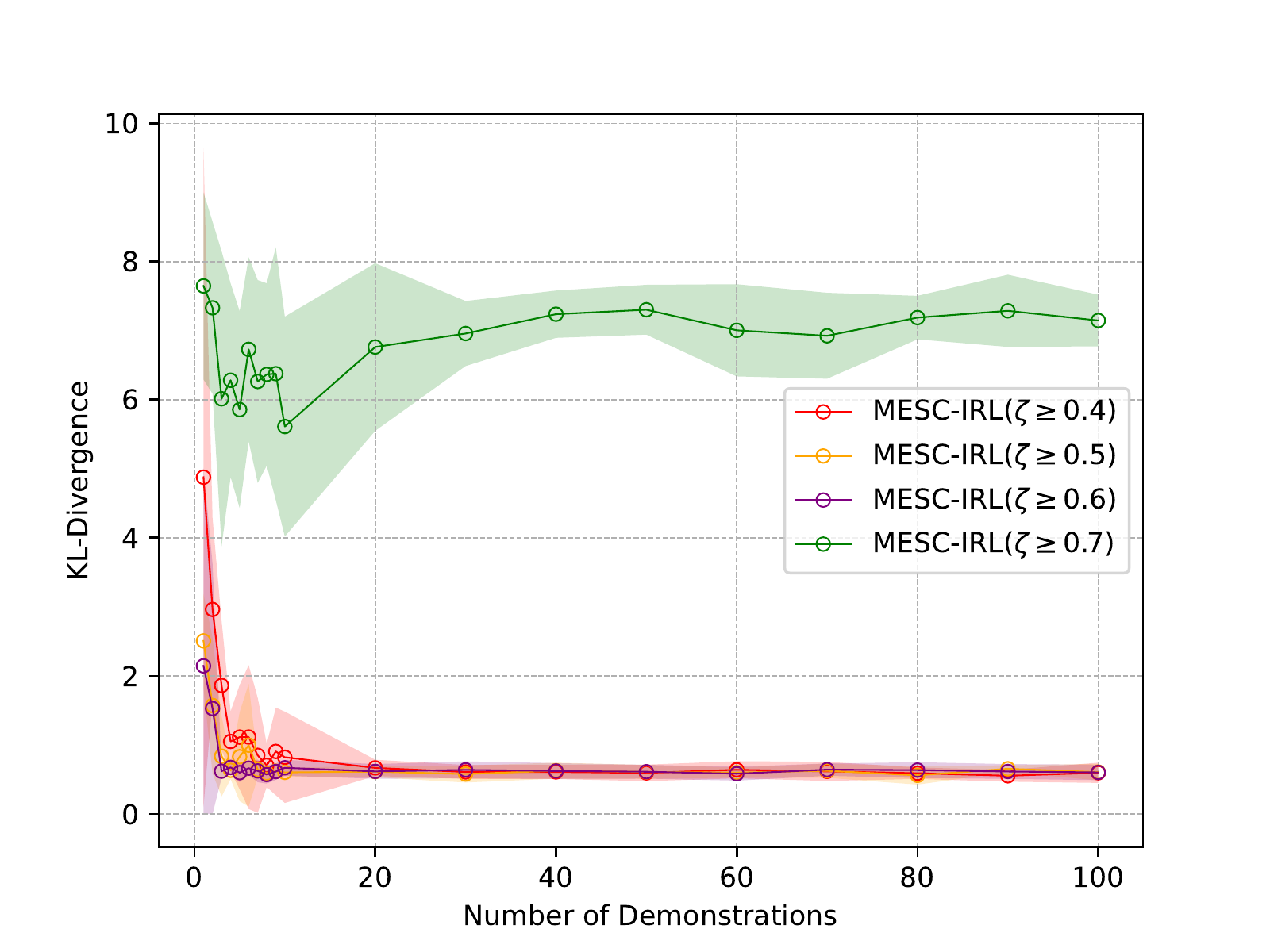}
\end{minipage}
\caption{Performance of MESC-IRL for various settings of $\zeta$ at recovering hard constraints in a deterministic setting according to false positive rate (left) and KL-Divergence from the demonstrations $\demonstrations$ (right) as we vary the number of demonstrations. Each point is the mean of 10 independent draws.}
\label{fig:stepping}
\end{figure*}

Additional graphs and results for comparision with the methods of \citet{scobee-2020-maximumlikelihood} and MESC-IRL. Figure \ref{fig:stepping} shows the stepping results for MESC-IRL.

\begin{figure*}[h]
\begin{minipage}[t]{.48\textwidth}
  \centering
  \includegraphics[width=\linewidth]{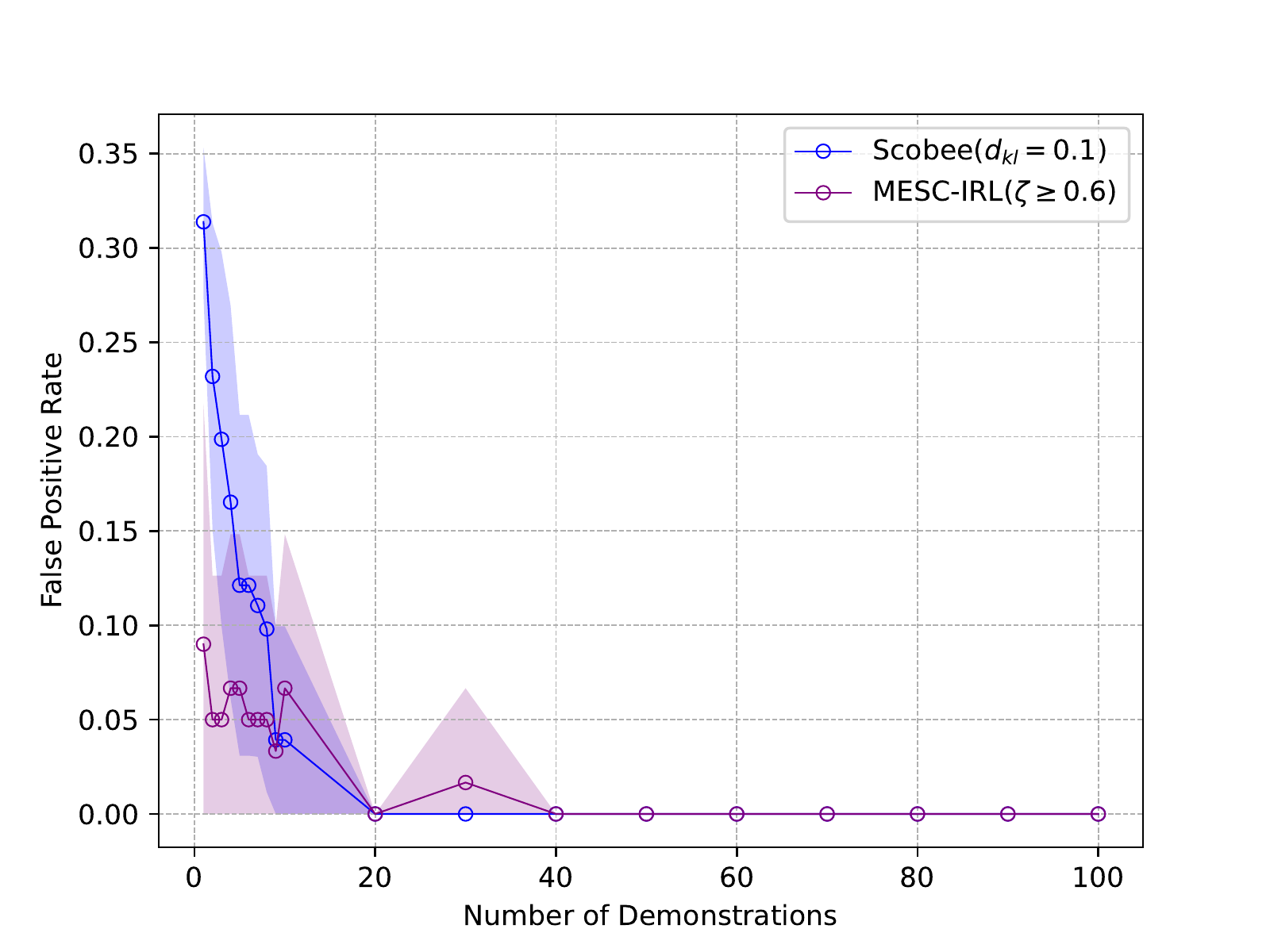}
\end{minipage}%
\hfill
\begin{minipage}[t]{.48\textwidth}
  \centering
  \includegraphics[width=\linewidth]{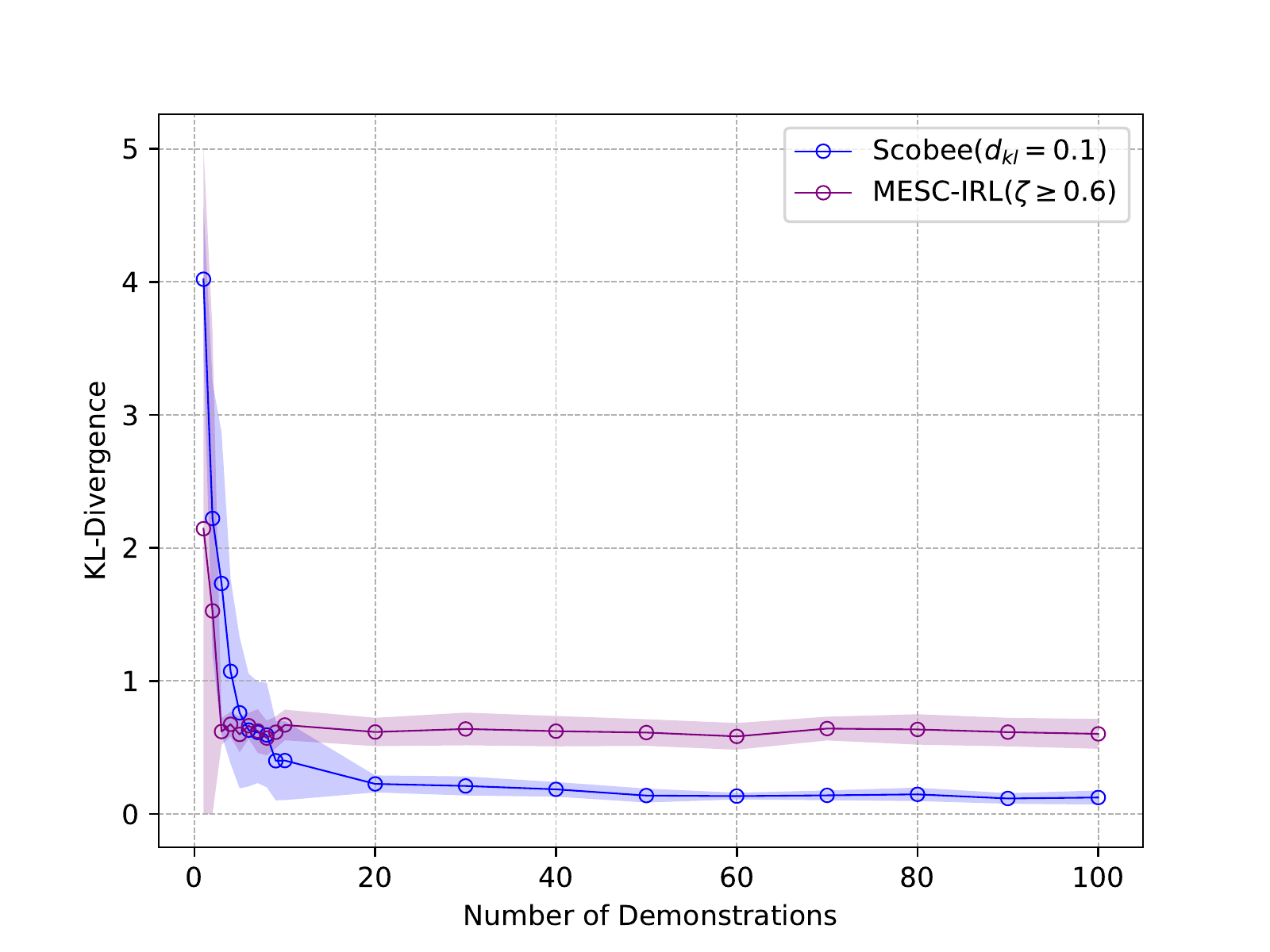}
\end{minipage}
\caption{Comparision of the best performing version of MESC-IRL to the best performing method of \citet{scobee-2020-maximumlikelihood} at recovering hard constraints in a deterministic setting according to false positive rate (left) and KL-Divergence from the demonstrations $\demonstrations$ (right) as we vary the number of demonstrations. Each point is the mean of 10 independent draws.}
\label{fig:scobee_hard_compare}
\end{figure*}

Figure \ref{fig:scobee_hard_compare} shows the performance of our best cutoff with the best method from \citet{scobee-2020-maximumlikelihood}.

\begin{figure*}[h]
\begin{minipage}[t]{.32\textwidth}
  \centering
  \includegraphics[width=\linewidth]{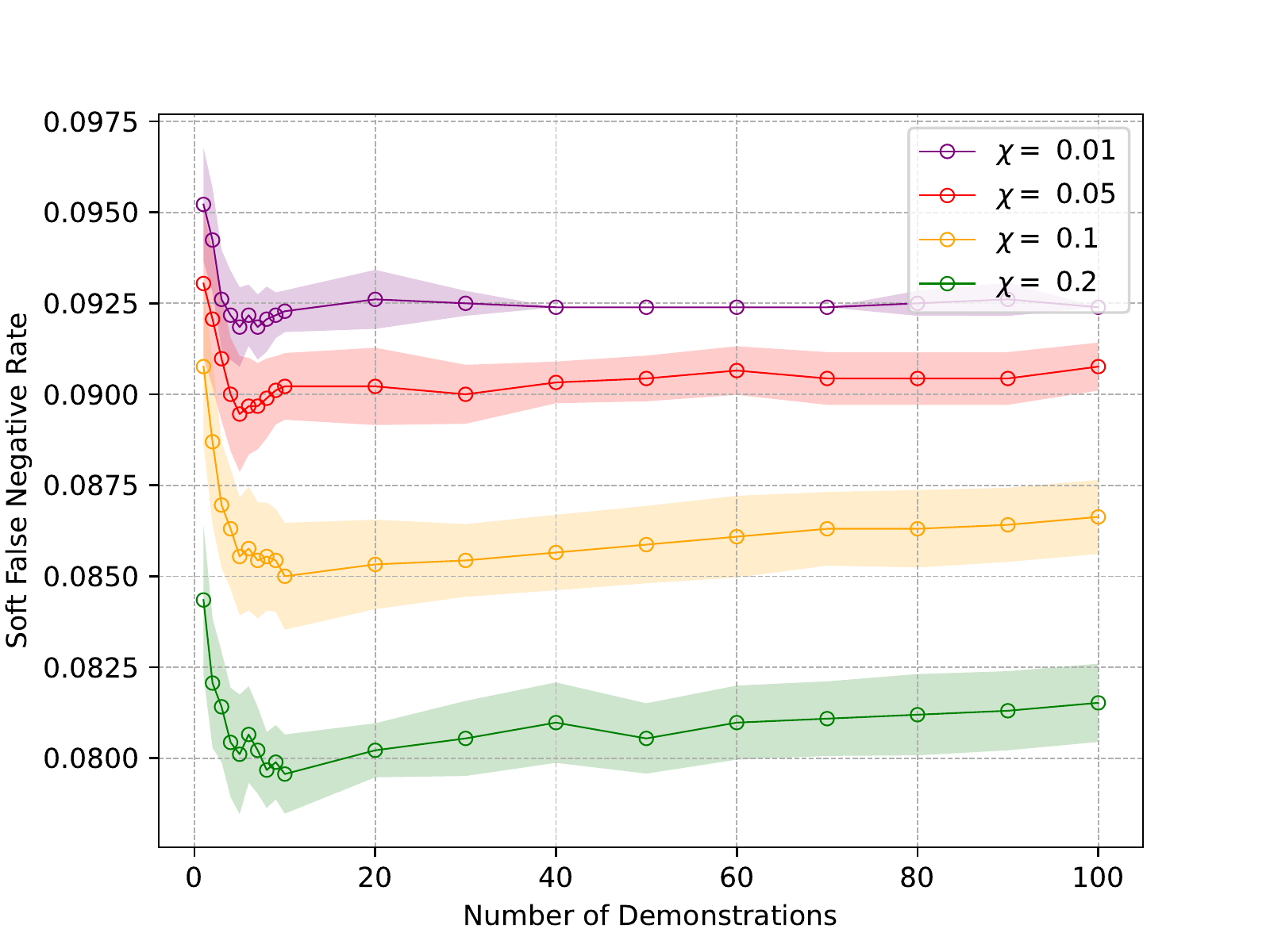}
\end{minipage}%
\hfill
\begin{minipage}[t]{.32\textwidth}
  \centering
  \includegraphics[width=\linewidth]{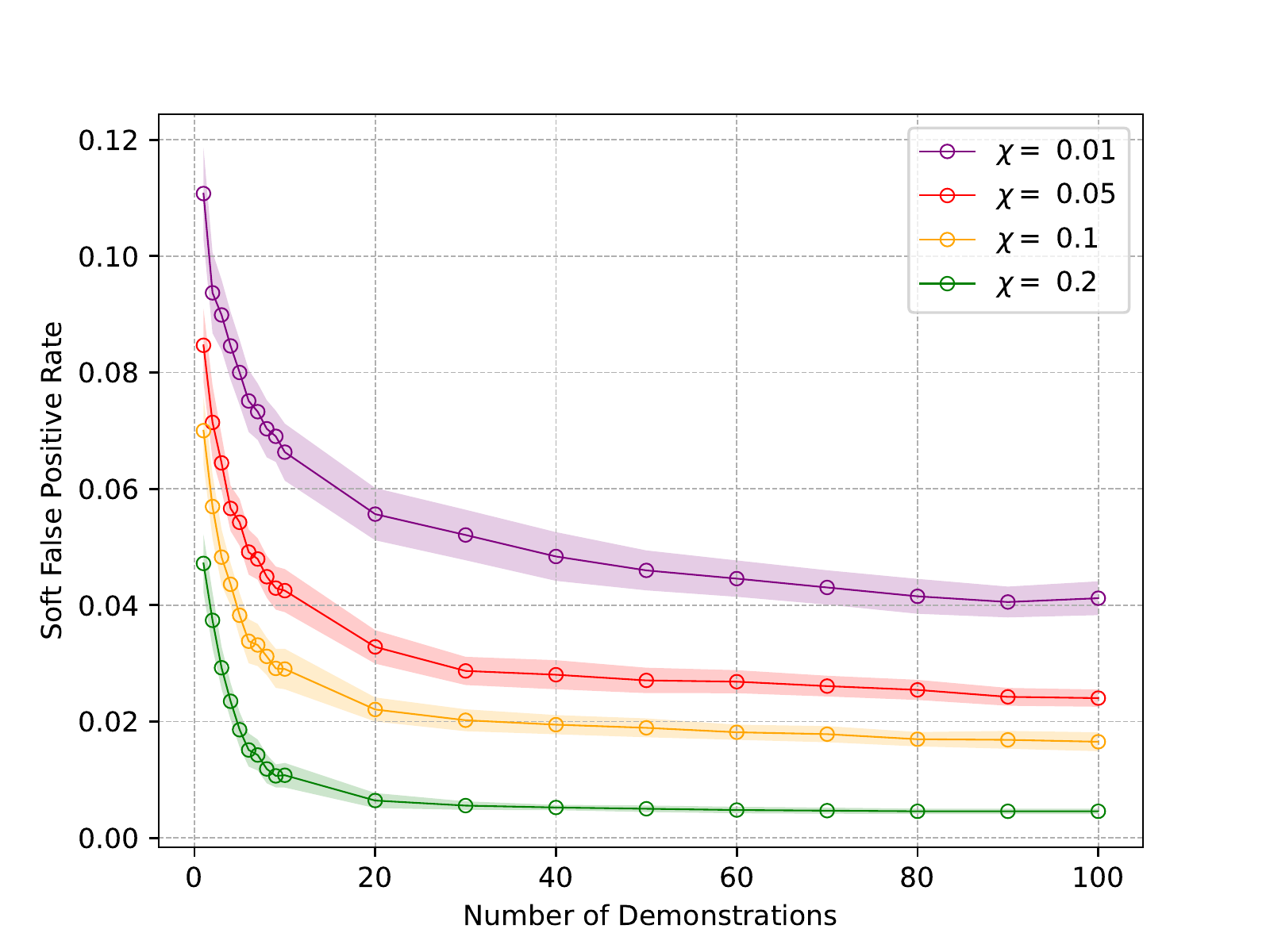}
\end{minipage}%
\hfill
\begin{minipage}[t]{.32\textwidth}
  \centering
  \includegraphics[width=\linewidth]{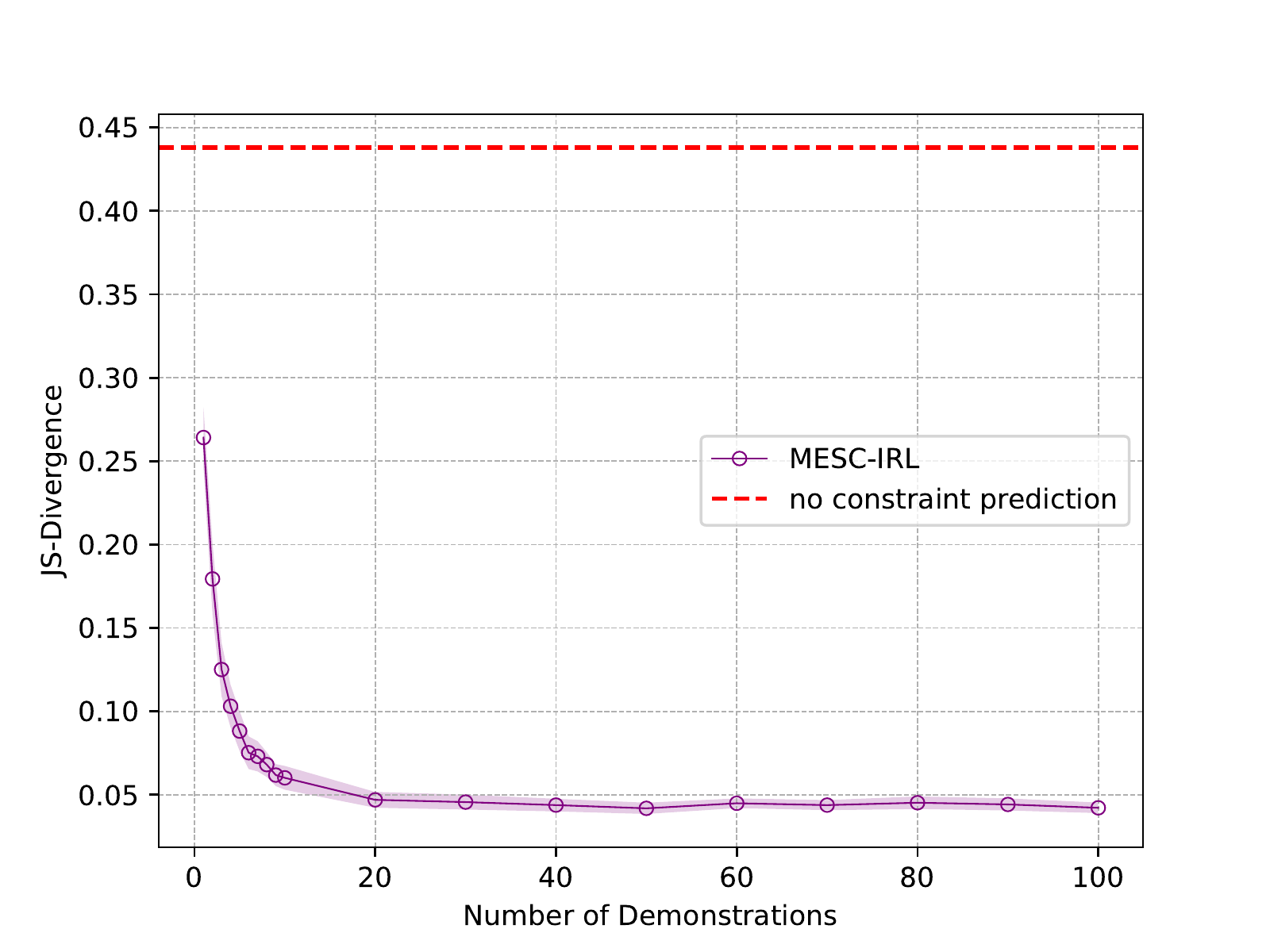}
\end{minipage}
\caption{Performance of MESC-IRL on recovering soft constraints in deterministic settings according to false negatives (left), false positives (center), and JS-Divergence to $\demonstrations$ (right). We see that across all these settings we are able to accurately recover constraints and generate behavior similar to the $\demonstrations$ even with few demonstrations.}
\label{fig:soft_deterministic}
\end{figure*}

\clearpage
Figure \ref{fig:soft_deterministic} shows the performance of MESC-IRL on recovering soft constraints in the deterministic setting.

\begin{figure*}[h]
\begin{minipage}[t]{.32\textwidth}
  \centering
  \includegraphics[width=\linewidth]{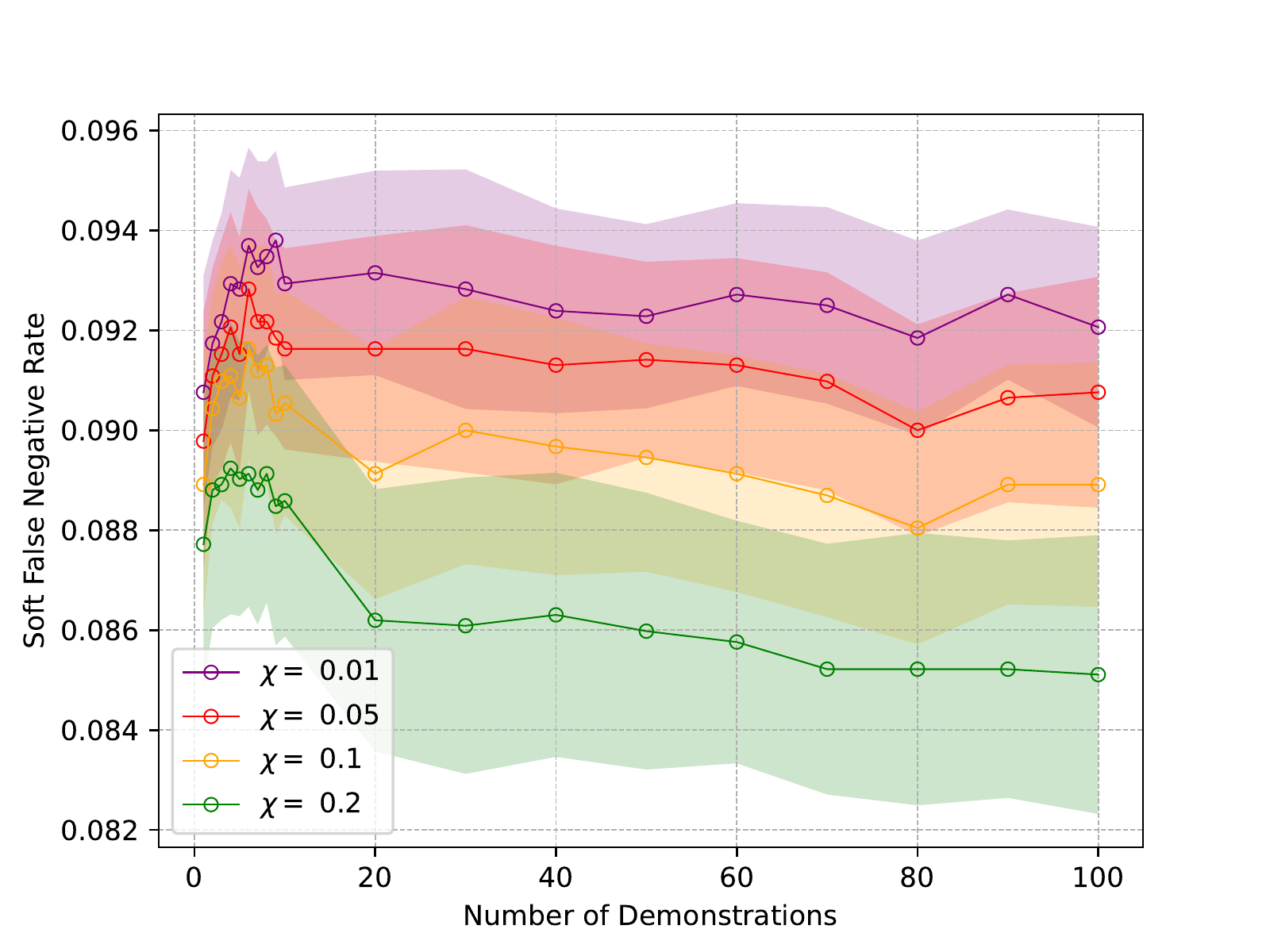}
\end{minipage}%
\hfill
\begin{minipage}[t]{.32\textwidth}
  \centering
  \includegraphics[width=\linewidth]{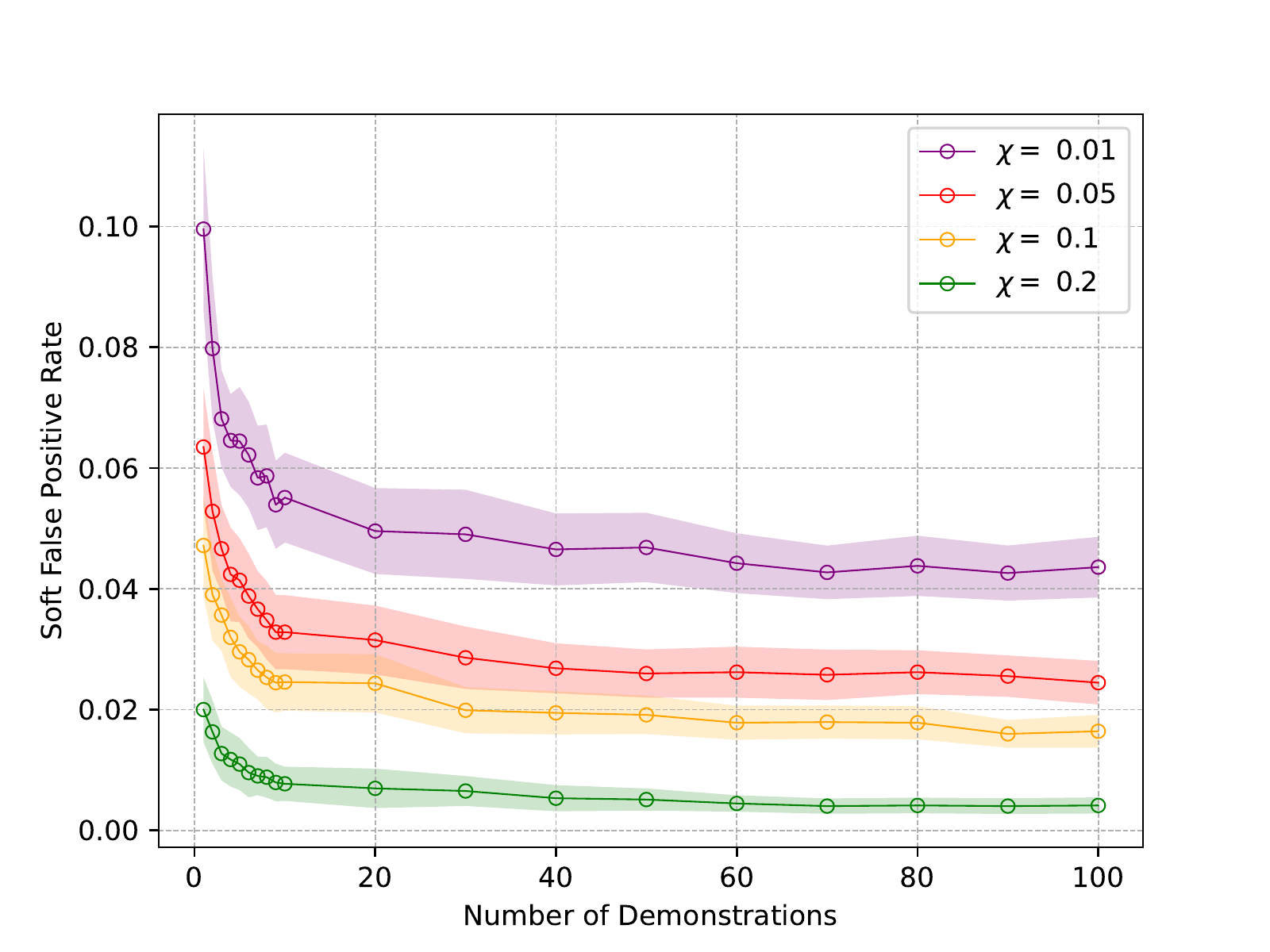}
\end{minipage}%
\hfill
\begin{minipage}[t]{.32\textwidth}
  \centering
  \includegraphics[width=\linewidth]{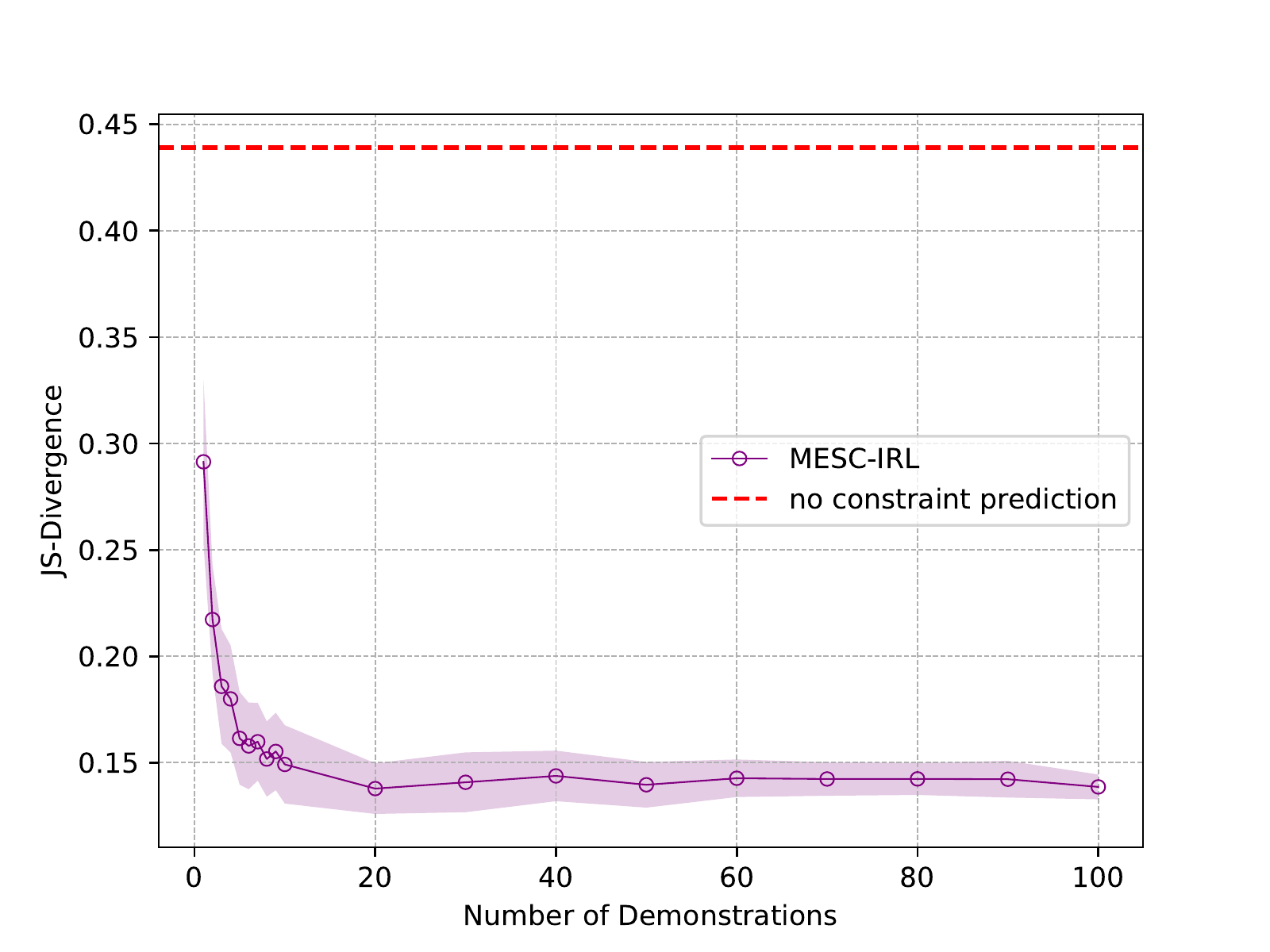}
\end{minipage}
\caption{Performance of MESC-IRL on recovering soft constraints in deterministic settings according to false negatives (left), false positives (center), and JS-Divergence to $\demonstrations$ (right). We see that across all these settings we are able to accurately recover constraints and generate behavior similar to the $\demonstrations$ even with few demonstrations.}
\label{fig:soft_nondet}
\end{figure*}

Figure \ref{fig:soft_nondet} shows the performance of MESC-IRL on recovering soft constraints in the non-deterministic setting.

\clearpage
\section{Proof Details for Comparison of Orchestration Methods}

We provide the full proofs for the theoretical comparison of the orchestrators.

%
%
%
%

\begin{theorem}
Consider  state $s$. Any choice probability distribution over the actions available in $s$ that can be modeled by the Greedy approach can be modeled via the MDFT or WA approaches. 
\end{theorem}

{\bf Proof.} We can model the (degenerate)  probability distribution induced by Greedy via an MDFT with as many options as the actions available in $s$,  two attributes with weights set to any random pair of values, and preferences in the $\mathbf{M}$ matrix all equally to 0 except for those in the row associated with $a$ which are set to 1. 
Matrices $\mathbf{C}$ and $\mathbf{S}$ can be defined in the standard way described in Section \ref{sec:mdft-back} and deliberation can be halted after one deliberation step.
In fact, when deliberation is launched, an attribute will be selected. Regardless of which one is selected, action $a$ will be chosen given that it is the only one with non-zero preference.

Similarly, we can model the Greedy distribution using a weighted average where $w_n=w_c=1/2$, and $sq_n(s,a)=sq_c(s,a)=1$ and $sq_n(s,a')=sq_c(s,a')=0$, $\forall a' \neq a$. $\Box$

\begin{theorem}
\label{mdft-gen}
Given state $s$ and the set $\mathcal{A}_s$ of actions available in $s$,
consider a probability distribution $p$ defined over $\mathcal{A}_s$.
We can define an MDFT model where the set of options corresponds to $\mathcal{A}_s$ and the induced choice probability distribution coincides with $p$.
\end{theorem}
{\bf Prrof.} Consider the MDFT model defined as follows:
\begin{itemize}
    \item Matrix $\mathbf{M}$ is the  $k \times k$ identity matrix;
    \item Weight vectors $\mathbf{W}$ are defined as in Section \ref{sec:mdft-back} and select a single attribute at each iteration. 
    Probability distribution over attributes $\mathbf{w}$ is defined in a way such that the probability of selecting the $j$-th attribute, is $p(a_j)$. 
  \item Matrices $\mathbf{C}$ and $\mathbf{S}$ are defined in the standard way as described in Section \ref{sec:mdft-back}.
    \item The deliberation time for the model is fixed at one iteration.
\end{itemize}  
It is easy to see that running the model induces a choice probability over the actions which corresponds to $p$. In fact, in every run, which consists of a single iteration, an attribute $A_h$ will be sampled according to probability $p$. Given how $\mathbf{M}$ is defined and the fact that the initial value of the accumulated preference $\mathbf{P}(0)=0$, action $a_h$ will be chosen.  Thus, 
the probability of  action $a_h$ being selected, given the MDFT model,  coincides with  $p(a_h)$. $\Box$

\end{document}